\documentclass[11pt, letterpaper]{article}
\usepackage[letterpaper, margin=1in]{geometry}
\usepackage[utf8]{inputenc}
\usepackage[style=numeric,backend=biber]{biblatex}
\usepackage[english]{babel}
\usepackage{csquotes}
\usepackage{hyperref}
\usepackage{booktabs} 
\usepackage[table]{xcolor} 
\usepackage{multirow} 
\usepackage{pdfpages} 
\usepackage{graphicx} 

\graphicspath{ {./images/} }
\usepackage[style=numeric,backend=biber]{biblatex}
\newcommand{\paymentUSD}{20}

\begin{document}
\noindent {\Large \bf Technical Report UTEP-CS-23-27}

\bigskip\bigskip\medskip

\noindent {\Huge \bf Dialogs Re-enacted Across Languages,}

\medskip
\noindent {\Huge \bf Version 2}

\bigskip\bigskip \noindent  Nigel G. Ward,  Jonathan E. Avila, Emilia Rivas, Divette Marco

\bigskip\noindent 
nigelward@acm.org, jonathan.edav@gmail.com, rivasemilia2@gmail.com, divettemarco@outlook.com

\bigskip
\noindent Department of Computer Science, University of Texas at El Paso

\bigskip

\noindent June 27, 2023  

\noindent \bigskip \bigskip

\noindent {\bf Summary}

\smallskip \noindent To support machine learning of cross-language
prosodic mappings and other ways to improve speech-to-speech
translation, we present a protocol for collecting closely matched
pairs of utterances across languages, a description of the resulting
data collection and its public release, and some observations and
musings.  This report is intended for:

\smallskip 
\begin{tabular}{l}
      \textbullet{} ~~ people using this corpus     \\
      \textbullet{} ~~ people extending this corpus \\
      \textbullet{} ~~ people designing similar collections of bilingual dialog data
\end{tabular}

\bigskip\noindent {\bf Change Notes} This version supersedes
UTEP-CS-22-108.  There is some new information and numerous
clarifications, mostly arising from our experiences diversifying our
corpus and helping a vendor to use this protocol.

\bigskip\noindent
{\bf Contents}
\smallskip

\begin{tabular}{ll}
      1   & Motivation                          \\
      2   & Related Work and Aims               \\
      3   & Strategy                            \\
      4   & Data Collection Procedure           \\
      &~~~~ overview, participants, equipment, step-by-step procedure, post-processing \\
      5   & Data Released to Date               \\
      6   & Observations                        \\
     &~~~~ participant behavior, data quality \\
      7   & Towards Increased Variety           \\
      8   & Cost Components                     \\
      9   & Other-Language Explorations         \\
      10  & An Exploration in Remote Collection \\
      \multicolumn{2}{l}{References} \\
      \multicolumn{2}{l}{Appendices} \\
\end{tabular}

\section{Motivation}

Future speech-to-speech translation systems should be able to effectively support
dialog, but current systems are not specifically built for this. One reason is the lack
of bilingual data on how dialog behaviors differ across languages. Thus there is a need
for parallel corpora of conversational speech across languages.
This report presents the
Dialogs Re-enacted Across Languages (DRAL) protocol and corpus.
The key idea is gathering
unscripted conversations among bilingual speakers who later re-enact selected utterances from
those conversations in their other language.

We hope that this data will be useful for many purposes, including at
least five: First, we ourselves are using these utterance pairs to
train models for mapping Spanish to English prosody, and conversely
\cite{avila-ward23}. For this we will try first modeling utterances or
phrases in isolation and then considering the local context.  Second,
this data could also support linguistic studies of the prosodic
differences between the two languages.  Third, it could support study
of other language phenomena, such as the meanings and translation of
dialog markers and discourse
connectives~\cite{luthier-popescu-belis-2020-chat}.  Fourth, this data
could be used to fine-tune translation models trained on large,
non-dialog data.  Finally, this data could be used to evaluate the
performance of different speech-to-speech translation systems for
dialog relevant scenarios.

We acknowledge that this data collection method is expensive and could never produce
corpora large enough to train lexical translation models or language models.  However
small, well-designed corpora can be very informative regarding many aspects of
conversational behavior~\cite{floyd2021conversation,dingemanse2014,dingemanse2022text}.

\section{Related work and corpus desiderata}

As already noted, an important use case for future speech-to-speech
translation systems is support for people in dialog, as they interact
across languages. While speech-to-speech translation systems so far
have been developed using mostly monologue speech data, participants
in dialog seldom speak in monologue style; rather their utterances are
often pragmatically rich, as the speakers take turns, shift topics,
take stances, try to amuse or persuade, and so on. Systems which are
blind to such intents will be of only limited value for supporting
model these behaviors across languages, but corpora of equivalent
spoken dialog data across languages have been lacking.

Classic corpora for speech translation research, as surveyed by~\cite{doi2021large},
have been largely designed to support research on speech-to-text translation. However the
development of true speech-to-speech systems requires corpora that include
target-language audio.

Recent years have seen the development of several true multilingual
speech corpora, often motivated by the need for data to train
end-to-end speech-to-speech translation
systems~\cite{jia2019direct,lee2021direct,zhang2020uwspeech,kano2021transformer}. Most
such corpora are exclusively monologue and feature mostly read speech,
such as short prompts, Wikipedia articles, books, and religious texts
\parencite{common-voice,wang2020a,pratap2020,boito-mass}. As reading
lacks interactive aspects, these corpora exhibit no dialog phenomena
and little pragmatic richness. Even monologue corpora which are not
just reading, for example political discussions and informative talks
\parencite{voxpopuli,cattoni2021must,salesky2021}, exhibit only modest
style variation. Parallel corpora derived from dubbed television shows
\parencite{oktem2021corpora,baali2022creating,huang2023holistic}, exhibit more
pragmatic variety, but these are scripted, and thus neither
spontaneous nor truly interactive, and are moreover performed by
professional actors and interpreters, and thus depart significantly
from natural conversational behavior. Some interesting corpora have
been described but are unfortunately not
available~\cite{doi2021large}. The corpora which come closest to
meeting the need are those which start with actual spoken
conversations and include target-language equivalents produced by
translating the dialog transcripts and then feeding the resulting
target-language text to a speech
synthesizer~\cite{zhang2020uwspeech,jia2022cvss}. While suitable for
some purposes, the synthesized outputs lack meaningful prosodic
variation, and generally fail to faithfully match the intent of the
original.

Thus, although useful in various ways, no existing resources support
development of speech-to-speech translation systems for dialog
scenarios. In particular, there is an unmet need for corpora that are
natural, faithful, exhibit wide pragmatic diversity, and are
available.

\section{Strategy}

For the sake of both pragmatic diversity and naturalness, we start by  recording
nonprofessional speakers in unstructured conversations.
We then collect re-enactments in another language.

For the sake of fidelity, we use bilingual speakers.
Thus each matched utterance pair
in our corpus was the product of a single speaker. This avoids the need at modeling time to design
mechanisms to disentangle differences due to speaker variation from differences due to
the language spoken.

Having each speaker re-enact their own previous utterances has the
further advantage of increasing the likelihood of faithfully recalling
and faithfully reproducing the exact intent and meanings of the
original utterance.

Achieving reasonable naturalness required some thought on how to obtain the
re-enactments. In theory we could have had participants re-enact an entire conversation
in one go, but this would have required them to transcribe and translate the words as a
memory aid, and likely to go through multiple passes of rehearsal, making the result
more like read speech and less spontaneous~\cite{wagner2016acted}. Therefore we chose to have them re-enact the
conversation fragment-by-fragment. While we considered having participants re-enact
utterances as they went along, turn by turn, this would have broken the flow of the
conversation. Thus we had them do the re-enactments immediately after the original
conversation finished.

While participants might, in theory, be able to manage everything themselves,
we chose to have each session moderated by a ``director,'' variously below also called the
``operator'' and ``producer,'' as one person performed all these roles.

Choosing the length of the fragments involved a trade-off: too long, and it would be
hard for the participants to retain the full intention in working memory; too short, and
the context would be lacking. We did not set any fixed target, but let the operator and
participants find what lengths worked well. In the data collected so far, the
re-enactments average  3.65 seconds.    

As the re-enactments are at times internally complex, sometimes
including contributions from both speakers and/or multiple utterances
from one speaker, we also release individual phrases, or ``short
fragments.''  Each of these includes audio from only one speaker and
includes only one phrase, and as such these pairs will likely be more
convenient for most modeling tasks.  
These phrases average 2.65 seconds. 

\section{Data collection procedure}

\subsection{Overview}\label{sec:overview}

To reiterate, the basic idea is that pairs of participants have a
short conversation in one language and then re-enact fragments of that
conversation in another language. An operator handles the recording
and guides the participants.

To help participants get in the mood, at each stage the operator instructs them in the
language they are about to be using.

Since overlap is common in spontaneous conversation, we chose to record the
speakers in separate channels, with the best audio separation we could reasonably
attain.  For this reason, we have participants wear head-mounted headsets, and sit in separate rooms.
The signals from the microphones are fed
into a stereo digital recorder and also played back to the headphones of the other speaker so they can hear each other.

The original conversations last about 10 minutes.  If they were longer, we would be gathering more data than we could ever re-enact.  If they were shorter, participants would not have time to get comfortable with each other and get into interesting topics. Also if they were shorter, we would at times lack enough interesting utterances to re-enact.
We also find that 10 minutes of free conversation puts them in a good mood and perhaps more willing to work
hard on the difficult task of re-enacting utterances.

\begin{figure}[h]
      \centering
     \includegraphics[scale=0.31]{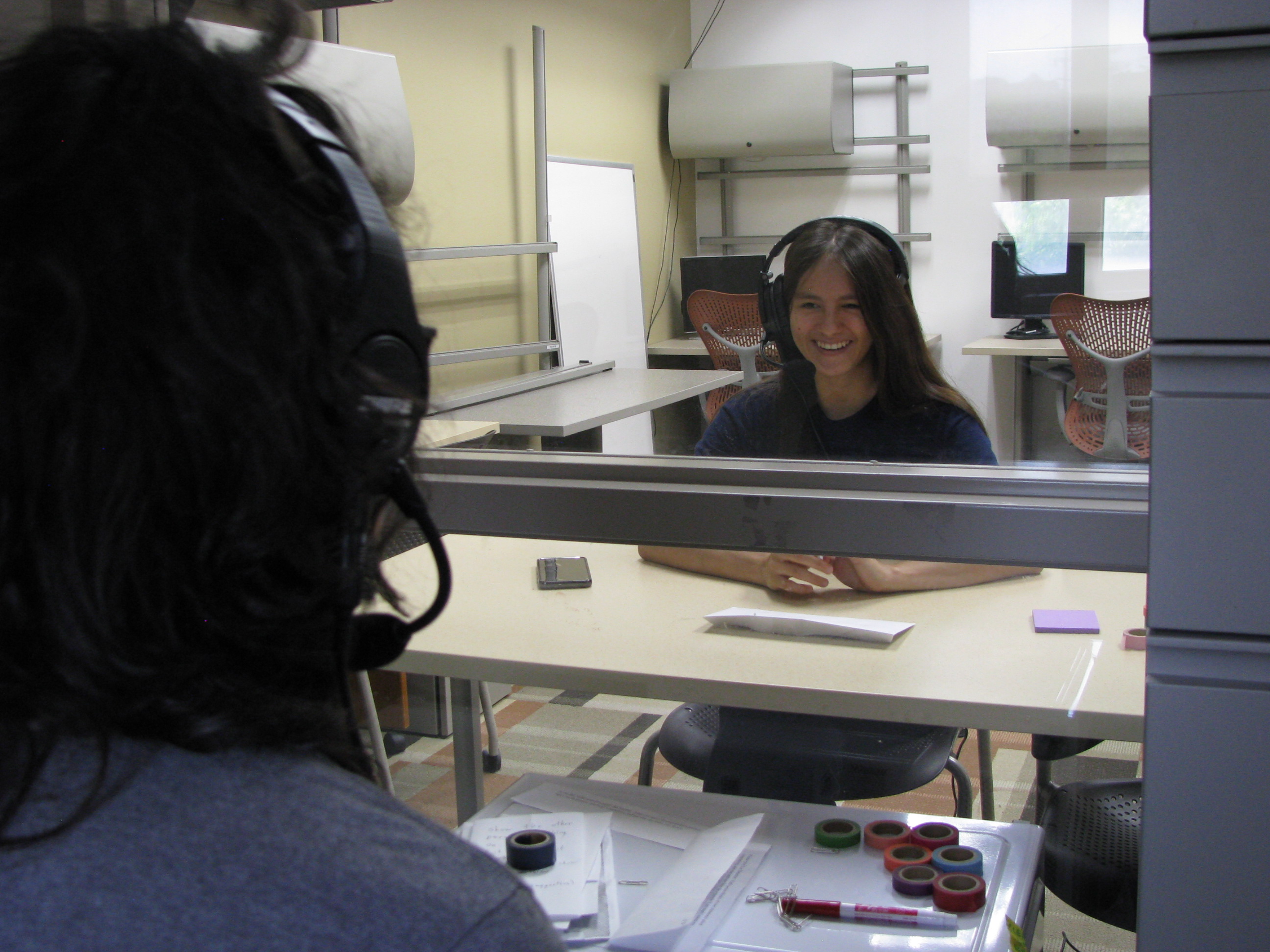}
\caption{Interacting  through the glass wall. } \label{fig:glass-wall}
\end{figure}

The original conversations are done with the participants in two rooms
which have audio separation but a glass window so they can see each
other (Figure \ref{fig:glass-wall}). We close the doors to the main
room not only to improve audio separation but to enable them to have a
private conversation, without feeling monitored or feeling pressure to
put on a performance.

\begin{figure}[h]
      \centering
      \includegraphics[scale=0.31]{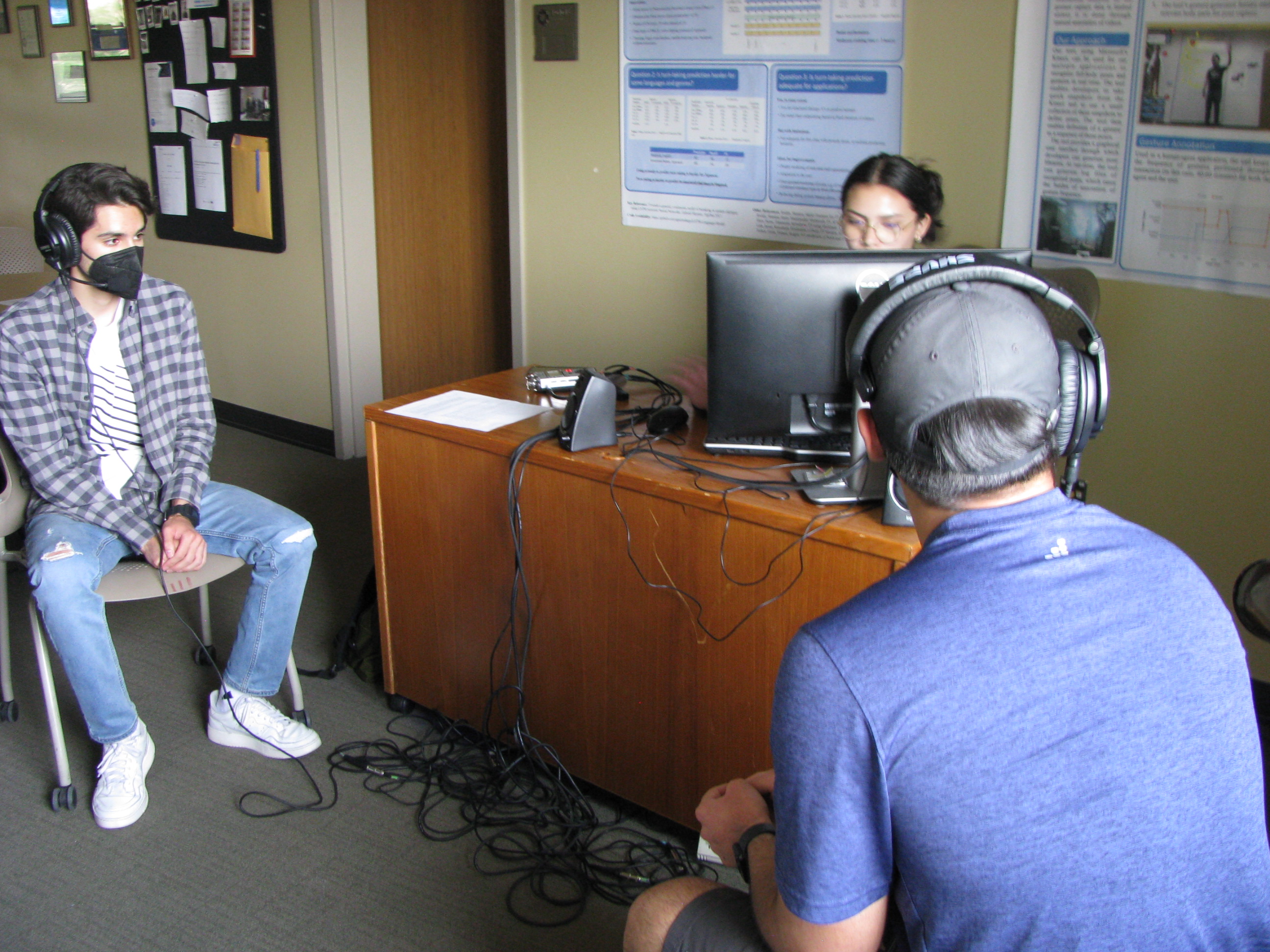}\caption{Re-enactment phase.} \label{fig:reenactment}
\end{figure}

For the re-enactment stage, the participants come into the main room,
so they can interact more easily with the operator, while still
maintaining some distance for the sake of retaining at least some
audio separation (Figure \ref{fig:reenactment}).  After identifying
any segments of the conversation that the participants want redacted,
the operator scrubs through the recording, using ELAN, and together
with the participants, chooses utterances or utterance sequences for
re-enactment.

Selecting utterances for re-enactment involves many factors.  Since we
do not have a specific use case in mind, our primary consideration is
``interestingness.''  Uninteresting utterances are those of types of
which we already have many in the corpus, such as initial formalities
and utterances from the midst of telling stories.  Thus the operator
preferentially chooses utterances that are interesting in the sense of
exhibiting behaviors or attitudes less commonly seen.  These are also
typically more interesting for the participants.  As a rule of thumb,
utterances in regions of active engagement, often marked by the
presence of laughter, tend to be more interesting.  Also, for the sake
of diversity of speakers, and to keep both participants engaged, when
one person dominated the original conversation, the operator attempts
to balance out the re-enactments across both speakers.

Another consideration in utterance selection is the turn-taking.
On the one hand, we generally avoid sequences with rapid back-and-forth turn-taking,
both because they are hard to replicate and because they are unlikely to occur in
any use case of conversations mediated by a system.
On the other hand, we often choose fragments that have speech from both participants. This
helped keep both participants engaged throughout the process. In most such fragments,
one participant's contribution was only a back-channel, laughter, or short response.
Such short utterances are very hard to produce realistically in isolation, so
re-enacting them in context helped improve their fidelity and naturalness. Further,
including these served to  encourage the main speaker to re-create the exact same feeling in
order to elicit the exactly equivalent response.

The operator instructs the participants to make their re-enactments ``feel the same
way'' as the original speech, allowing for different word choice.
The consent form uses the wording ``re-create the mood and feeling of each utterance as closely as possible''
and the operator reinforces this idea many times, in different words.  The participants are also
encouraged to re-enact overlaps, disfluencies and pauses as much as possible.

Each selected fragment is played repeatedly for the participants,
until they indicate that they are ready to re-create it. The
participant or participants translate and re-enact the utterance or
utterance sequence. Re-enactments occasionally require multiple
attempts to attain enough naturalness and fidelity to satisfy the
participants and the operator. When five attempts are not enough, that
fragment is abandoned.  There is something of a trade-off here, and we
might explore whether to limit play back of each fragment to just a
couple of repetitions before re-enactment, to reduce the pressure to,
and opportunity to, mimic it too faithfully, in order to increase the
chances of obtaining equally natural but more distinct target-language
forms.

To avoid fatiguing the participants, sessions last no more than one hour.
For this reason also, not all utterances are re-enacted.
At the end, the participants are asked again
whether they consent to share the recordings. Finally, each is paid; this was 15 USD and is now  \paymentUSD{} USD.\@

Later, the operator uses his or her handwritten notes on the start
times of the re-enacted fragments to use ELAN to mark up the exact
spans and cross-identify them with the corresponding spans in the
original conversation. In a second pass, the operator marks individual
phrases (short fragments) within both the original and the re-enacted
utterances, typically pause-separated, and cross-identifies them.
Finally, we run a script to pull out the matching phrases from the
recordings, as described in Section~\ref{post-processing-procedure}.

\subsection{Participants}

We recruit participants from among students at the University of Texas at El Paso, which
is located on the US-Mexico border.
Many students have family members on both sides of the
border, many students have received part of their education on both sides, and many
commute across it daily. Even on the US side, where the dominant language is
English, many neighborhoods are Spanish-dominant.

We advertised for Spanish-English bilinguals. Our advertisement is in
Appendix~\ref{appendix:advertisement}. Their self-described dialects
were, for English, overwhelmingly El Paso and, for Spanish, mostly
``El Paso / Juarez.'' Other dialects included East Texas,  Chihuahua City,
 Guanajuato, and Ojinaga.

We encouraged participants to bring a partner to talk with, but if
they came alone, the operator served as their conversation
partner. Both of our operators grew up in El Paso and are native and
fluent in both languages.

\subsection{Hardware and room configuration}

\begin{figure}[h]
      \centering
      \includegraphics[scale=0.5]{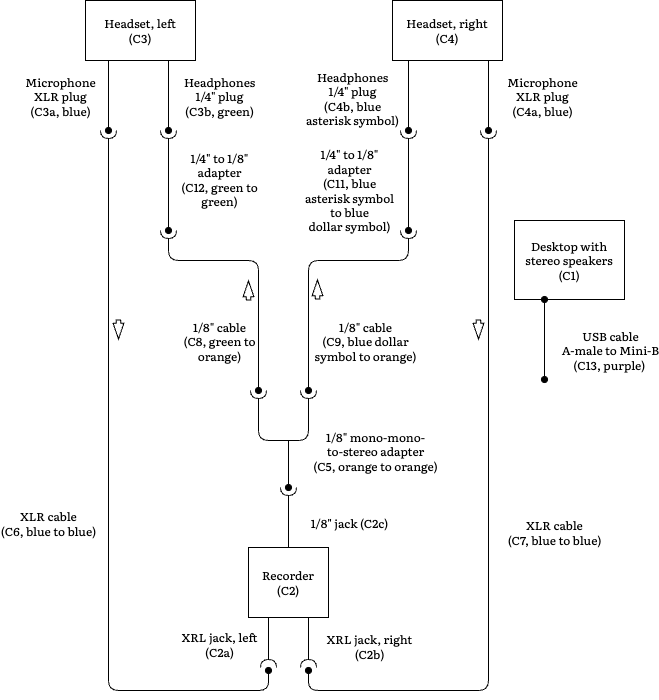}\caption{Hardware
            diagram. Component numbers and color labels are in parentheses.}\label{figure:hardware}
\end{figure}

The goal of the hardware setup is to record a conversation such that
interlocutors are recorded on separate audio tracks, while still being
able to hear each other. Our setup (Figure~\ref{figure:hardware})
includes two identical single-sided headsets (Shure BRH441M) and a
digital recorder with microphone inputs (TASCAM DR-40).  Audio is
recorded in WAV format, sampling at 44.1 kHz. Working with the cables
we had, we added cable adapters where needed. These adapters should
not have a significant effect on quality.

\begin{figure}[h]
      \centering
      \includegraphics[scale=0.5]{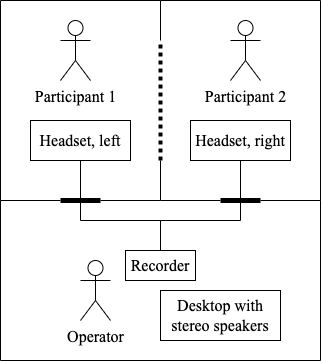}\caption{Configuration of rooms.
            The dotted line is the glass window and the thick lines are doors.  }\label{figure:room}
\end{figure}

The data collection takes place in a research lab with a main room connected to two
separate rooms with a glass window between them, as suggested by Figure~\ref{figure:room}.

\subsection{Data collection instructions}\label{collection-procedure-details}

\subsubsection{Setting up for the initial
      recording}\label{setting-up-for-initial-recording}

\begin{enumerate}
      \item Take inventory of the hardware:
            \begin{itemize}
                  \item one desktop computer with stereo speakers (C1)
                  \item one recorder (C2)
                  \item two headsets (C3, C4)
                  \item two microphone cables (shielded, thick, three-prong) (C6, C7)
                  \item two audio cables (thin) (C8, C9)
                  \item one USB upload cable (C13)
                  \item three adapters (C5, C11, C12)
            \end{itemize}
            The desktop with stereo speakers (C1) is next to the bookshelf. On the bookshelf, the box labeled ``TASCAM DR-40'' contains (C2) and the ``Under Armour'' shoebox  contains (C3) through (C13). Refer to Figures~\ref{figure:hardware} and~\ref{figure:room} when connecting components.
      \item Take the components out of the boxes. Remove the Velcro and twist ties and keep them to store the components later. Some components are labeled ``L'' (left) and ``R'' (right).
      \item Place the recorder (C2) and cable (C13) in the main room, within reach. The USB upload cable (C13) will not be used until the next stage.
      \item Place a chair in each of the two smaller rooms, near the glass window and facing each other.
      \item Place the left headset (C3) on the chair in the left room.
      \item Place the right headset (C4) on the chair in the right room.
      \item Connect the left headset microphone to the recorder.
            \begin{enumerate}
                  \item Connect plug (C3a) to cable (C6).
                  \item Connect cable (C6) to jack (C2a).
            \end{enumerate}
      \item Connect the right headset microphone to the recorder.
            \begin{enumerate}
                  \item Connect plug (C4a) to cable (C7).
                  \item Connect cable (C7) to jack (C2b).
            \end{enumerate}
      \item Connect the left headset headphones to the mono-mono-to-stereo adapter.
            \begin{enumerate}
                  \item Connect plug (C3b) to adapter (C12).
                  \item Connect adapter (C12) to cable (C8).
                  \item Connect cable (C8) to adapter (C5).
            \end{enumerate}
      \item Connect the right headset headphones to the mono-mono-to-stereo adapter.
            \begin{enumerate}
                  \item Connect plug (C4b) to cable (C9).
                  \item Connect cable (C9) to adapter (C11).
                  \item Connect adapter (C11) to adapter (C5).
            \end{enumerate}
      \item Connect the mono-mono-to-stereo adapter (C5) to the recorder (C2c). Leave this step to last, because the weight of the cables can damage the recorder.
      \item Lay all the cables flat and separated, so they do not catch when the doors are closed.
      \item Turn on the recorder by pressing and holding the HOME button on the front of the recorder. If the recorder does not turn on, check that the HOLD switch on the side of the recorder is not engaged.
      \item Sound-check the headsets. Press the RECORD button on the front of the recorder, then snap your fingers or speak close to each microphone in turn and check the recorder display. The bars on the display stretch to the right if the microphones are picking up sound. The top bar corresponds to the left microphone and the bottom bar corresponds to the right microphone.
      \item Turn on the desktop to have it ready for the next stages. To unlock the desktop, press CTRL + ALT + Delete.
      \item Sound-check the desktop speakers. Play a stereo audio, e.g.\ from a YouTube video, and check that each speaker is outputting sound.
      \item From the green folder in the desk, take out a blank piece of paper to write notes in the next stage.  Look  up the most recent assigned subject numbers, and prepare to assign the sequential next numbers to the new participants.
\end{enumerate}

\subsubsection{When participants arrive}\label{when-participants-arrive}

\begin{enumerate}
      \item Welcome participants into the main room and have them sit at the round table
            to go over consent forms and any questions. Let them know that, as they read
            on the flyer, there will be a lot of translating from one language to the
            other, and it is crucial to the study that they are fluent in both, for example, by saying:
            \begin{displayquote}
                  Before we begin and sign anything, I'm instructed to inform you that
                  there will be a lot of translating from one language to the other, so
                  if you feel like if that is something you won't be able to do, then I
                  wouldn't want to waste your time. Otherwise, we can proceed.
            \end{displayquote}
      \item Casually speak in both languages to gauge how fluent they are in each. Ask
            them to practice translating with you.
            \begin{itemize}
                  \item If both participants are fluent, continue with the procedure.
                  \item If only one of the participants is fluent, add a note of this
                        and continue with the procedure. We may still use the audio.
                  \item If both of the participants are not fluent, inform the
                        participants that you have ``gathered what we needed for the
                        study'' then give them the \paymentUSD{} USD and dismiss them.
            \end{itemize}
      \item Give each participant a consent form (Appendix~\ref{appendix:consent-form})
            and a language background form (Appendix~\ref{appendix:background-form}),
            found in the bookshelf.
      \item When participants hand you their filled-out forms, enter the participants'
            assigned participant ID on their language background form: the same
            participant IDs you assigned before they arrived.

          \item Go over things one more time, or as many times as
            they need in order to understand the purpose and
            procedure. For example, you might say:
            \begin{displayquote}
              The purpose of this data collection is to further speech-to-speech
                  translation research by creating an open collection of translated
                  conversations, something that has not been done before.

                  Today you will have a conversation with your partner in one language,
                  then re-enact parts of it in another language. I will select some
                  snippets of the audio and replay them for you to translate re-record
                  in the other language. It is important that you try your best to make
                  it sound natural while also keeping the same feeling as in the
                  original. Try to recreate pauses, laughs, long breaths, or anything of
                  that sort during the second recording if possible.

                  I can replay the audio as many times as you need, and give you as much
                  time as you need to translate. If either of us feel like you can
                  translate the words better or if the prosody was not as faithful
                  in feeling as it could be, then we can redo it until we are satisfied. Please be vocal of any
                  opinions that you have about the process and ask any questions that
                  may arise.
            \end{displayquote}
      \item Optionally suggest a topic or dialog activity (Section~\ref{sec:increasing-variety}).
      \item Ask the participants to move into their separate rooms and put on their
            headsets.
      \item Press the RECORD button on the recorder again. Make sure
        it is recording. 
      \item While the participants are talking, on the blank note sheet, record the
            following information: the current date, the ID of the conversation
            (following the format from Table~\ref{table:conversation}),
            the left-track (left room)  participant ID, and the right track (right
            room) participant ID. If you notice that they have participated before, refer back to a previous "unique ID" that you may have assigned for them, as this will be used to identify new versus  previously-unseen participants. \@
      \item After 10 minutes, let the participants know that the
        recording is over, and now they will begin translating and
        re-recording fragments of the conversation.  Ask them to wait
        a couple of minutes while you set up.

\end{enumerate}

\subsubsection{Setting up for the re-enactment
      recording}\label{setting-up-for-reenactment-recording}

\begin{enumerate}
      \item Disconnect the recorder (C2) from the other components (C5, C6, C7). To disconnect (C6, C7), press and hold the release tabs next to the jacks (C2a, C2b).
      \item Turn on the desktop speakers by raising the volume with the volume knob.
      \item Connect the recorder (C2) to the desktop (C1) using cable (C13).
      \item Press the HOME button on the front of the recorder. From the options that
            appear on the recorder display, select STORAGE.\@ If the menu does not
            appear, disconnect and reconnect cable (C13), then press the HOME button again.
      \item When the recorder storage window appears on the desktop, open the MUSIC
            folder.
      \item Copy the most recent WAV file to the \texttt{IsgAudioRecordings} folder on
            the desktop.
      \item Rename the audio file with its conversation ID, following the format from
            Table~\ref{table:conversation}.
      \item Eject the recorder from the desktop, then disconnect cable (C13) from the recorder (C2) and desktop (C1).
      \item Reconnect the recorder (C3) to the other components (C5, C6, C7), as they were before.
      \item Open ELAN.\@ In the blank window that appears, select ``File'' \(>\) ``New''.
            Navigate to to the \texttt{IsgAudioRecordings} folder, select the WAV file
            you copied, and select ``OK'' to finish.
      \item An ELAN window should open with the audio on the bottom half and an
            annotation tier named ``default''.
      \item Set up the annotation tiers for the next stage.
            \begin{enumerate}
                  \item Rename the ``Default'' to ``Utterance''.
                  \item Add a new tier named ``LittleLeft''.
                  \item Add a new tier named ``LittleRight''.
            \end{enumerate}
\end{enumerate}

\subsubsection{Recording re-enactments}\label{translating-rerecording}
\begin{enumerate}
      \item Ask the participants if there are any parts of their conversation
            that they would like to have removed. Annotate these sections with
            ``DELETE'' (see Appendix~\ref{appendix:elan-keyboard-shortcuts} for Elan keyboard
            shortcuts).
      \item Emphasize that they should keep the same  feeling in their re-enactments.
      \item Turn on the recorder and press the RECORD button.\@
      \item If at any point during recording the re-enactments you notice that one or both of
            the participants are not as fluent as expected, follow the instruction in
            Step 2 of Section~\ref{when-participants-arrive}.
      \item Play the audio for the participants and you to hear, stopping at a
            fragment to re-enact, preferably one that is easy to understand and differs from
            the rest of the recording in terms of prosody or feeling.
            (What makes a ``good'' fragment to re-enact was discussed in Section~\ref{sec:overview}.)
      \item Mark this region the Utterance tier, starting with \#1 and incrementing.
      \item Replay this audio fragment at least two or three times, or more if the
            participants request.
      \item When the participant(s) succeed in producing a good re-enactment, jot down
            the time on your notes sheet, to reference in the next stage.
      \item Repeat Steps 5 -- 8  for the remainder of the session, leaving some time to
            debrief and pay the participants.
      \item Press the RECORD button to stop recording re-enactments.
      \item Ask the participants if they have any questions. If everything is clear, ask
            them to sign the re-consent forms. After they sign the re-consent forms, thank them for their participation, give
            them each \paymentUSD{} USD, and dismiss them.
\end{enumerate}

\subsubsection{When participants leave}
If there will be another recording session soon, set up the hardware again like in Section~\ref{setting-up-for-initial-recording}. If not, disconnect all the components and store them as they were found. Wrap cables carefully using the same velcro and twist ties. Turn off the speakers completely (until the volume knob clicks).

\subsection{Entering metadata and adding annotations}\label{metadata-entry}
\begin{enumerate}
      \item On the desktop, open the metadata Excel workbook in the
            ``IsgDatabaseTables'' folder.
      \item If this is your first time as producer, enter two new rows in the
            ``producer'' sheet. Table~\ref{table:producer} describes each field in the ``producer'' sheet.
      \item In the ``participant'' sheet, enter two new rows.
            Table~\ref{table:participant} describes each field in the ``participant'' sheet. Note: Each participant is assigned two IDs. For more information, see Table~\ref{table:participant}.
      \item Go over things one more time, or
      \item In the ``conversation'' sheet, enter a new row.
            Table~\ref{table:conversation} describes each field in the ``conversation'' sheet.
\end{enumerate}

Now you can start mapping fragments to their translations.

\begin{enumerate}
      \item Create a new ELAN file for the re-enacted conversation and configure the three annotation tiers, as done in Section~\ref{setting-up-for-reenactment-recording} Step 12.
      \item Open the ELAN file for the original conversation, so that you have both the ELAN file for the original conversation and the re-enacted conversation open, side-by-side. The original will already have utterances labeled. Typically, these will be integers starting with 1 and going up to 30 or so.
      \item For each fragment in the ELAN file for the original conversation, annotate the re-enactment in the re-enactment file with the same number as the fragment it recreates. Use the times that you noted in Section~\ref{translating-rerecording} Step 8to find the re-enactments quickly. See Appendix~\ref{appendix:elan-keyboard-shortcuts} for keyboard shortcuts.
      \item In addition, when possible or necessary, identify matching phrases within each original-re-enactment pair.
            This is done for two reasons: first, the prosody within a single phrase is expected to be simpler to
            model than the prosody of concatenated phrases, and second, we want these matched pairs to contain speech from only one participant.
            There are three cases.
            In the first case, there are re-enactments which already include speech from only one participant. If this is one inseparable unit, then we simply annotate the same range in the tier for that speaker, LittleLeft for the left speaker, and LittleRight for the right speaker. If however, there is a clear pause (or occasionally more than one), we mark the individual phrases as separate regions in the appropriate Little tier.
            The second case is for re-enactments that include speaker changes.  For these we mark the appropriate regions in the Little tiers.   For example, if the left participant talks then the right participant, the first region will be marked in LittleLeft and the second in LittleRight.
            The third case are re-enactments that involve overlap.  Again, we mark all speech regions for the left participant in LittleLeft and for the right in LittleRight.  In all three cases, we do this
            first for the original audio and then in the same way for the re-enactment audio.
      \item Save both files before closing the windows.
\end{enumerate}

\subsection{Post-processing procedure}\label{post-processing-procedure}

The script for creating a DRAL release reads the conversation audio,
markup, and metadata files, and then outputs: phrase pairs (also
called ``short fragments''), re-enactment pairs (also called ``long
fragments''), short fragment audios concatenated per conversation and
track (useful for computing normalization parameters), a copy of the
complete original and re-enacted audio recordings, and updated
metadata as CSV files.

\subsubsection{Set up}\label{post-processing-setup}

\begin{enumerate}
      \item Create a root directory for the project somewhere, e.g.
            \textasciitilde\texttt{/Desktop/DRAL}.
      \item Download the files from the GitHub repository
            \url{https://github.com/joneavila/DRAL} into the project
            root directory.
      \item Install SoX. To install SoX on macOS or Linux, see:
            \url{https://github.com/rabitt/pysox#install}. If installing on Windows, add the directory
            containing the SoX executable to the PATH environment variable.
      \item Install Python 3.10.4 or create a Python 3.10.4 virtual environment for the
            project.
      \item Install the required Python modules and their dependencies with pip:
            \texttt{pip install -r requirements.txt}. The main Python modules are:
            \textit{pandas} for data manipulation, \textit{pympi} for processing ELAN
            files, and \textit{pysox}, a Python SoX wrapper.
      \item Copy the directory \texttt{recordings/}, containing the conversation
            audio (\texttt{.wav}) and markup (\texttt{.eaf}) files, to the project root
            directory.
      \item Copy the metadata worksheet \texttt{metadata.xlsx} to the project root
            directory.
\end{enumerate}

\subsubsection{Creating a new release}\label{creating-a-release}

The script \texttt{make\_release.py} expects that the audio, markup, and metadata
files are in the same parent directory, as described in
Section~\ref{post-processing-setup}. It writes the contents of the release to the output
directory \texttt{release/}. To create a new release, simply run the script:
\texttt{python3 make\_release.py}. If the command fails, see the output help message for a summary of the expected arguments.

The script will print warnings for problematic conversations and
fragments. These are skipped and not included in \texttt{release/}
unless the problems are fixed and the script re-run. Each warning will
hint at a possible fix (Table~\ref{table:script-warnings}).

\bigskip

\begin{table}[ht]
    \begin{center}
      \caption{Post-processing script warnings and causes.}\label{table:script-warnings}
      \begin{tabular}{ p{5cm} p{10cm} }
            \toprule
            Warning                                                                 &
            Cause                                                                                                                                            \\
            \toprule
            These conversations are missing markup\dots                             &
            The conversation does not have a markup file with the same name as its ID.\@
            \\
            \toprule
            These conversations are missing audio\dots                              &
            The conversation does not have an audio file with the same name as its ID.\@
            \\
            \toprule
            These conversations have an unexpected ID\dots                          &
            The conversation does not have an ID following the pattern: a valid language
            code, followed by an underscore, followed by three digits.                                                                                       \\
            \toprule
            These conversations have an unexpected original or re-enacted code\dots &
            The conversation does not have an original or re-enacted code following the
            pattern: a value of ``OG'' or ``RE''.                                                                                                            \\
            \toprule
            These conversations do not have a translation with a valid ID and original
            and re-enacted code or have multiple translations\dots
                                                                                    & The conversation does not have exactly one translation with an ID that
            matches the pattern: a valid language code other than the language code in
            the ID of the current conversation, followed by an underscore, followed by
            the same three digits in the ID of the current conversation.                                                                                     \\
            \toprule
            These markups have unexpected values\dots                               &
            The markup has a value that does not match the pattern: an optional pound
            symbol, followed by any number of digits. The markup value might contain a
            typo or contain a comment to be excluded from the release.                                                                                       \\
            \toprule
            These markups are in an unexpected tier\dots                            &
            The markup is in a tier with a name that does not follow the pattern: a
            value of ``LittleLeft'' or ``LittleRight'' or ``Utterance''.                                                                                     \\
            \toprule
            These markups have duplicate values\dots                                &
            The markup shares a value with one or more markups in the same tier (the
            markup value is not unique).                                                                                                                     \\
            \toprule
            These markups have zero or more than one translation\dots               &
            The fragment does not have exactly one translation with a markup value equal
            to the current fragment and in a tier with the same name as the current
            fragment.                                                                                                                                        \\
            \toprule
      \end{tabular}
      \end{center}
\end{table}

\subsubsection{Quality control}
We do not have a systematic quality control process. Periodically
three of us (project leader, data consumer and workflow designer,
operator) sit down for an hour to listen to the matching pairs from a
few dialogs.  We comment on interesting cases and boring cases, and
note cases which lapsed somewhat in fidelity or, more rarely,
naturalness.  In the rare case that incorrect pairings are found, as
may result from errors in the markup stage, we mark these for
re-annotation and thus correction in the next release. More commonly,
we discuss what went well or less well.  Outcomes have included
adjusting the balance of instructions to prioritize naturalness in the
target language and coming up with ways to diversify the dialog acts
present.

\section{Data Released to Date}

We make this corpus public.  As noted above, we anticipate its uses to
include research and training and evaluating speech-to-speech
translation models and systems. 

Our data collection is ongoing, and we are releasing it as we go.
Table~\ref{table:corpus-stats} summarizes the status, through DRAL 7.0, that is, as of June 27, 2023.  These numbers do not include our held-out test data, derived from another 32 conversation pairs, which we are, for now, keeping pristine for possible use in a future shared task. 
Those interested in this or other uses are welcome to contact any of the authors. 

Releases include the short fragments (individual phrases), which we
expect will be the primary unit of analysis for most purposes, and
also the recordings of the entire original conversations and the
entire re-enactment sessions. We also release all scripts, so that
others can reprocess the data as convenient for their purposes. We
also release the metadata, notably the ID numbers for the participants
in each dialog, cross-referenced to their dialect information, etc.
Tables~\ref{table:participant},~\ref{table:conversation},
and~\ref{table:fragments} describe the columns in the CSV files.

Releases are available at \url{www.cs.utep.edu/nigel/dral/}, which is
updated periodically.  Each additional release contains additional
conversation recordings and their phrases along with revised,
cumulative metadata for the entire corpus.  When extraction errors are
discovered, they are fixed in the next release, in the form of revised
audio files to replace the old ones. Thus, to get the entire corpus,
download each installment, starting with DRAL-2.0, and unpack each into
its own folder.  Discard all metadata (csv files) except those from
the last release.  Create directories for recordings/,
fragments-long/, and fragments-short/ , then populate each by copying
over all files from the corresponding directories in the releases.  Do
this copying starting with 2.0 and working forward, to ensure that
any revised audio files overwrite the old ones.  

\begin{table}[th]
      \begin{center}
            \caption{Corpus statistics . }\label{table:corpus-stats}
            \begin{tabular}{l r}
                  \toprule
                  Conversations              & 104 pairs    \\
                  \toprule
                  Unique Participants               & 70    \\
                  \toprule
                  Matched re-enactment pairs & 2263   \\
                  \toprule
                  Mean re-enactment duration & 3.65~s \\
                  \toprule
                  Matched phrase pairs       & 2893  \\
                  \toprule
                  Mean phrase duration       & 2.65~s \\
                  \toprule
            \end{tabular}
      \end{center}
\end{table}

\begin{table}[ht]
      \caption{Participant table fields. All fields are entered
            manually.}\label{table:participant}
      \begin{tabular}{p{3cm} p{6cm} p{6cm}} Field                                 & Description
                                                                            & Value                                                                                                                                                                                                                                                                                              \\
               \toprule
               \textit{id}                                                  & A unique identifier for a participant.
                                                                            & A sequence of integers.
               \\
               \textit{id\_unique}                                          & A second identifier for the participant. This field is used to distinguish repeat participants from first-time participants. & A sequence of integers. First-time participants are assigned a unique value, whereas repeat participants are assigned the same value assigned in previous sessions. \\
               \textit{lang1}                                               & The first of the two languages the participant
               speaks (in either the original or re-enacted conversation).  & An ISO 639--1 two-letter code, e.g. ``EN'' for English, ``ES'' for Spanish.
               \\
               \textit{lang2}                                               & The second of the two languages the participant
               speaks (in either the original or re-enacted conversation).  & An ISO 639--1 two-letter code, e.g. ``EN'' for English, ``ES'' for Spanish.
               \\
               \textit{lang\_strength}                                      & The participant's strengths in
               language 1 and language 2, self-reported in a questionnaire. & An integer value 1
               through 5, representing a spectrum: 1 ``language 1 stronger'', 2
               ``language 1 slightly stronger'', 3 ``equal'', 4 ``language 2 slightly
               stronger'', and 5 ``language 2 stronger''.                                                                                                                                                                                                                                                                                                                        \\
               \textit{dialect\_note1}                                      & The note on the participant's language
               1 dialect, self-reported in a questionnaire.                 & A string, e.g. ``El Paso''.
               \\
               \textit{dialect\_note2}                                      & The note on the participant's language
               2 dialect, self-reported in a questionnaire.                 & A string, e.g. ``Chihuahua City''.
               \\
               \textit{is\_producer}                                        & Whether the participant is also a
               producer.                                                    & A boolean: asterisk (``*'') for True, blank for False.
               \\
               \toprule
      \end{tabular}
\end{table}

\begin{table}[ht]
      \caption{Conversation table fields. The field \textit{trans\_id} is entered
            automatically by the post-processing script.}\label{table:conversation}
      \begin{tabular}{p{3.1cm} p{6cm} p{6cm}}
            Field                                            & Description
                                                             & Value
            \\
            \toprule
            \textit{id}                                      & A unique identifier for a conversation,
            matching the filename of the conversation audio. & A string with
            format: \(<\)language code\(>\_<\)three digits\(>\), e.g. ``ES\_018''.                                                                                                                                          \\
            \textit{recording\_date}                         & The date of the recording session.
                                                             & A date string: dd/mm/yyyy.
            \\
            \textit{original\_or\_reenacted}                 & Whether the conversation is original or the re-enacted.
                                                             & ``OG'' for original or ``RE'' for reenacted.
            \\
            \textit{participant\_id\_left}                   & The unique identifier of the participant
            recorded on the left track.                      & The same value entered in Table~\ref{table:participant} \textit{id} field.                                                                                   \\
            \textit{participant\_id\_right}                  & The unique identifier of the participant
            recorded on the right track.                     & The same value entered in Table~\ref{table:participant} \textit{id} field.                                                                                   \\
            \textit{participant\_id\_\-left\_unique}         & The second identifier of the participant
            recorded on the left track.                      & The same value entered in Table~\ref{table:participant} \textit{id\_unique} field.                                                                           \\
            \textit{participant\_id\_\-right\_unique}        & The second identifier of the participant
            recorded on the right track.                     & The same value entered in Table~\ref{table:participant} \textit{id\_unique} field.                                                                           \\
            \textit{producer\_id}                            & The unique identifier of the producer.                                             & The same value entered in Table~\ref{table:producer} \textit{id} field. \\
            \textit{trans\_id}                               & The unique identifier of the translation conversation.                             & A value entered in Table~\ref{table:conversation} \textit{id} field.    \\
            \toprule
      \end{tabular}
\end{table}

\begin{table}[ht]
      \caption{Short fragments table and long fragments table fields. All fields are
            entered automatically by the post-processing script.}\label{table:fragments}
      \begin{tabular}{p{3cm} p{6cm} p{6cm}} Field              & Description
                                                         & Value
               \\
               \toprule
               \textit{id}                               & A unique identifier for a conversation fragment (utterance), matching the utterance's annotation value. & A string with format: \(<\)language code\(>\_<\)\textit{conv\_id}\(>\_<\)markup value\(>\), e.g. ``ES\_018\_1''. \\
               \textit{participant\_id}                  & (Short fragments only.) The unique identifier of the participant in the utterance.                      & A value entered in Table~\ref{table:participant} \textit{id} field.                                              \\
               \textit{participant\_id\_\-unique}        & (Short fragments only.) The second identifier of the participant
               in the utterance.                         & A value entered in Table~\ref{table:participant} \textit{id\_unique} field.                                                                                                                                                \\
               \textit{participant\_id\_left}            & (Long fragments only.) The unique identifier of the participant recorded on the left track.             & A value entered in Table~\ref{table:participant} \textit{id} field.                                              \\
               \textit{participant\_id\_right}           & (Long fragments only.) The unique identifier of the participant recorded on the right track.            & A value entered in Table~\ref{table:participant} \textit{id} field.                                              \\
               \textit{participant\_id\_\-left\_unique}  & (Long fragments only.) The second identifier of the participant
               recorded on the left track.               & A value entered in Table~\ref{table:participant} \textit{id\_unique} field.                                                                                                                                                \\
               \textit{participant\_id\_\-right\_unique} & (Long fragments only.) The second identifier of the participant
               recorded on the right track.              & A value entered in Table~\ref{table:participant} \textit{id\_unique} field.                                                                                                                                                \\
               \textit{lang\_code}                       & The language of the utterance's source conversation.                                                    & A value entered in Table~\ref{table:conversation} \textit{lang\_code} field.                                     \\
               \textit{conv\_id}                         & The unique identifier of the utterance's source conversation.                                           & A value entered in Table~\ref{table:conversation} \textit{id} field.                                             \\
               \textit{original\_or\_\-reenacted}        & Whether the utterance's source conversation is original or the re-enacted.                              & A value entered in Table~\ref{table:conversation} \textit{original\_or\_reenacted} field.                        \\
               \textit{time\_start}                      & The time into the source conversation the
               utterance starts.                         & A duration string with format: mm:ss.ms.
               \\
               \textit{time\_end}                        & The time into the source conversation the
               utterance ends.                           & A duration string with format: mm:ss.ms.
               \\
               \textit{duration}                         & The duration of the utterance.
                                                         & A duration string with format: mm:ss.ms
               \\
               \textit{trans\_id}                        & The unique identifier of the utterance's matching,
               translation utterance.                    & A value entered in Table~\ref{table:fragments} \textit{id} field.                                                                                                                                                          \\
               \toprule
      \end{tabular}
\end{table}

\begin{table}[ht]
      \caption{Producer table fields. All fields are entered
            manually.}\label{table:producer}
      \begin{tabular}{p{3cm} p{6cm} p{6cm}} Field & Description
                                            & Value                                                             \\
               \toprule
               \textit{id}                  & A unique identifier for a producer.                             &
               A sequence of integers.                                                                          \\
               \textit{name}                & The name of the producer. This field is excluded from releases. &
               A string.                                                                                        \\
               \toprule
      \end{tabular}
\end{table}

\section{Observations}

We ran the data collection protocol for over a year at our site, and
it worked well throughout.  We also supported a vendor as they adapted
our protocol for their context; they were also able to successfully
run and re-enact 27 conversations, annotate them, and process them
using our tools to produce a total of 589 matched pairs.  So we have
ample reason to believe that our protocol is solid.  The rest of this
section describes some observations, based on our own experience,
regarding the process and  the quality of the result.

\subsection{Participant Behavior} 
The original conversations seemed very natural and the participants
generally seemed to enjoy them; indeed sometimes it was hard to get
them to stop. While not all participants were equally fluent in both
languages, code switching was rare, mostly occurring when a
participant had a lexical gap, for example not knowing how to say
\textit{3D printer} in Spanish. 

 The re-enactment process was much more effortful, but our
 participants generally seemed to enjoy the challenges involved.  They
 understood that the purpose of our data collection was better
 translation systems, and they seemed eager to support it, setting
 high standards for the fidelity and naturalness of their productions,
 encouraged in this by the operator and probably some peer pressure.

Some participants enjoyed the process so much that they became
regulars, coming in repeatedly over the semester, sometimes bringing
different friends, although we note that this happened only after we
raised the compensation to \$\paymentUSD{}.  Some participants were
frequently able to produce high-quality re-enactments the first time;
others required more tries.  Some participants were able to smoothly
re-enact long utterances; others needed them broken down into shorter pieces. 

\subsection{Data Quality} 

In terms of quality, many of the original-conversation utterances were
disfluency-ridden, partly unintelligible, or strange from a grammatical, semantic, or
pragmatic viewpoint.  This is to be expected, as our participants were not professional
speakers.  However the utterances were generally  successful in terms of
interacting appropriately with the interlocutor and keeping the conversation going.

Participants appeared to vary in how they interpreted our instructions
regarding re-enactment.  Many appeared to strive for (or relax into)
lexical fidelity, with almost every word in the original represented
by a single word in the other, as in
Appendix~\ref{appendix:example-spontaneous} (but see also
Appendix~\ref{appendix:example-mismatch} as a partial counterexample).
Some seemed to strive for prosodic fidelity, producing re-enactments
where the overall prosodic contour was very similar to that of the
original.  Relatively early we honed our instructions to stress our desire
not for low-level fidelity but rather fidelity in terms of the feeling
and function conveyed.

The topics and interaction styles varied.  We noticed that some
participants deliberately avoided possibly contentious or embarrassing
topics (Appendix~\ref{appendix:example-spontaneous}), or seemed to
gravitate to ``safe'' topics or a low-effort speaking style, such as
telling stories or asking questions and giving answers.  These
tendencies worked against our aim of obtaining diverse pragmatic
functions, so we tried various ways to increase the diversity of
topics and dialog activities, as discussed in
Section~\ref{sec:increasing-variety}.

We wanted the re-enactments to be natural, and in this we largely
succeeded.  In informal experiments presenting pairs of original and
their re-enactments to bilinguals, we found they were mostly unable to
identify which was which, indicating that most of the re-enactments
were highly natural.  One common exception was utterances involving
laughter, which seems often very hard to re-create convincingly.  
Accordingly, we generally did not try to elicit re-enactments of laughter
for its own sake, but only for the reasons discussed above. 

In terms of functional diversity, after collecting the first dozen dialogs
we found that we already had at least  29
of the 42 dialog acts in Jurafsky {\it et al.}'s  (\citeyear{swbd-damsl}) list. In
addition, we observed many functions and forms not listed in such taxonomies, such as
saliently leaving something unsaid (Appendix~\ref{appendix:example-prosody}), strongly
asserting oneself (Appendix~\ref{appendix:example-spontaneous}), enacting reported
speech, and expressing sympathy.  In terms of Ward's dimensions of interaction style~\cite{ward2022using}, at various times in these dialogs there appeared to be interactions at each extreme of all 8 dimensions.

We observed that often much of the information in these utterances was
conveyed largely by the prosody (Appendix A.2, A.3).  In terms of
prosodic diversity, we observed examples of at least the 14 most
frequent prosodic constructions of English in Ward's
(\citeyear{ward-book}) listing. As interesting prosody is largely
lacking from existing speech-to-speech translation corpora, the
presence of many samples with meaningful and varied prosody is a
unique strength of this corpus.  We also did a preliminary scan over
short fragment pairs to judge the extent to which utterances with
identical intent would differ in prosody between English and Spanish
utterances. Somewhat surprisingly, given the literature on this
topic~\cite{bowen1956,ward-gallardo}, most often the differences were
minor, most saliently involving lexical stress and utterance-initial
and utterance-final lengthening, but see \cite{avila-ward23}. This may reflect properties of our
local dialects or perhaps the comfortable conversational styles that most participants
used.

\section{Towards Increased Variety}\label{sec:increasing-variety}

Although our data collection protocol already gives fairly good coverage
of the dialog acts and activities common in casual conversation, these
do not exhaust the interesting things that can happen in dialog.
Since many use cases involve dialog activities that we're
unlikely to observe by just inviting the participants to talk about
whatever they like, we have worked to increase the diversity in the
data by occasionally giving them suggestions.  These
are always optional.  While most participants willingly played
along, in most cases they quickly reverted to a normal conversation 
on their own preferred topics. 

Below we list some of the prompts, activities, and situations,
interleaved with notes on their purposes and observations on how these
affected the participants' behavior.

\begin{enumerate}
\item Agree on 3 bits of advice to give all new students joining the CS department (or coming to UTEP in general).

  This was to foster the activities of making plans together,
  including making suggestions, filtering them, and reaching
  agreement.

\item Think of a bad outcome that happened to you or someone you know, and together discuss who might share part of the blame, and who deserves most of the blame.

This and the previous are about discussing social norms and establishing responsibility.

\item	Ask the other person to show you everything in their purse or backpack.  Be persistent.

The intent was to elicit persuasion, requests, and negotiation, but
this led mostly to just explanations and stories.

\item	Talk about something that's on your mind, for example an intimidating homework assignment.

The intent was to foster  self-disclosure, and probably being supportive.

\item  Find something that the two of your disagree about.

The intent was to elicit opinions, self-reflection, and discussion of
contentious topics.

\item  Talk about something sad.

The intent was to broaden the emotional diversity of the corpus.  It also served to 
 broaden the topics. 

\item Here are two toy animals; each of you please hold one and come up with good names for them.

This helped the participants get engaged in the conversation
quicker. It led them to relate the toy to themselves in some way which
enabled the conversation to branch out into interesting or uncommon
topics. Brainstorming for names made participants talk about
individuals in their lives with those names. The toys led to
participants talking about entertainment from their childhood.

\item Please play Lie Detector. Each person thinks of two statements about themselves, one true and one false, for example `I went dancing last  week' and `I usually drink Red Bull every morning and afternoon.' After that, please talk about anything for the
rest of the 10 minutes.

Some started out with this game, but it was abandoned after a minute
or two in favor of topics that had been mentioned while playing the
game.

\item Tell each other  what actions your would take if you were a dictator of a country.

  The conversations were a mix of serious answers about how they would create positive change and jokey answers about how they would change the world to their liking.

  \item In the rooms you'll find several rolls of various colors of
    washi tape; try making an art piece on the glass window between
    you.

Some pairs did not use the tape at all.  In others, although the tape
was used while they talked, this did not contribute to the
conversation since the focus was on their topic. Other participants
just they played around with the tape during the recording. Though
in these cases it did not contribute to the conversation much, it still
provided the participants with something to do while conversing, which
could have made the participants more comfortable during the recording
as it gave them an escape from constant eye contact.

\begin{figure}[h]
      \centering
      \includegraphics[scale=0.31]{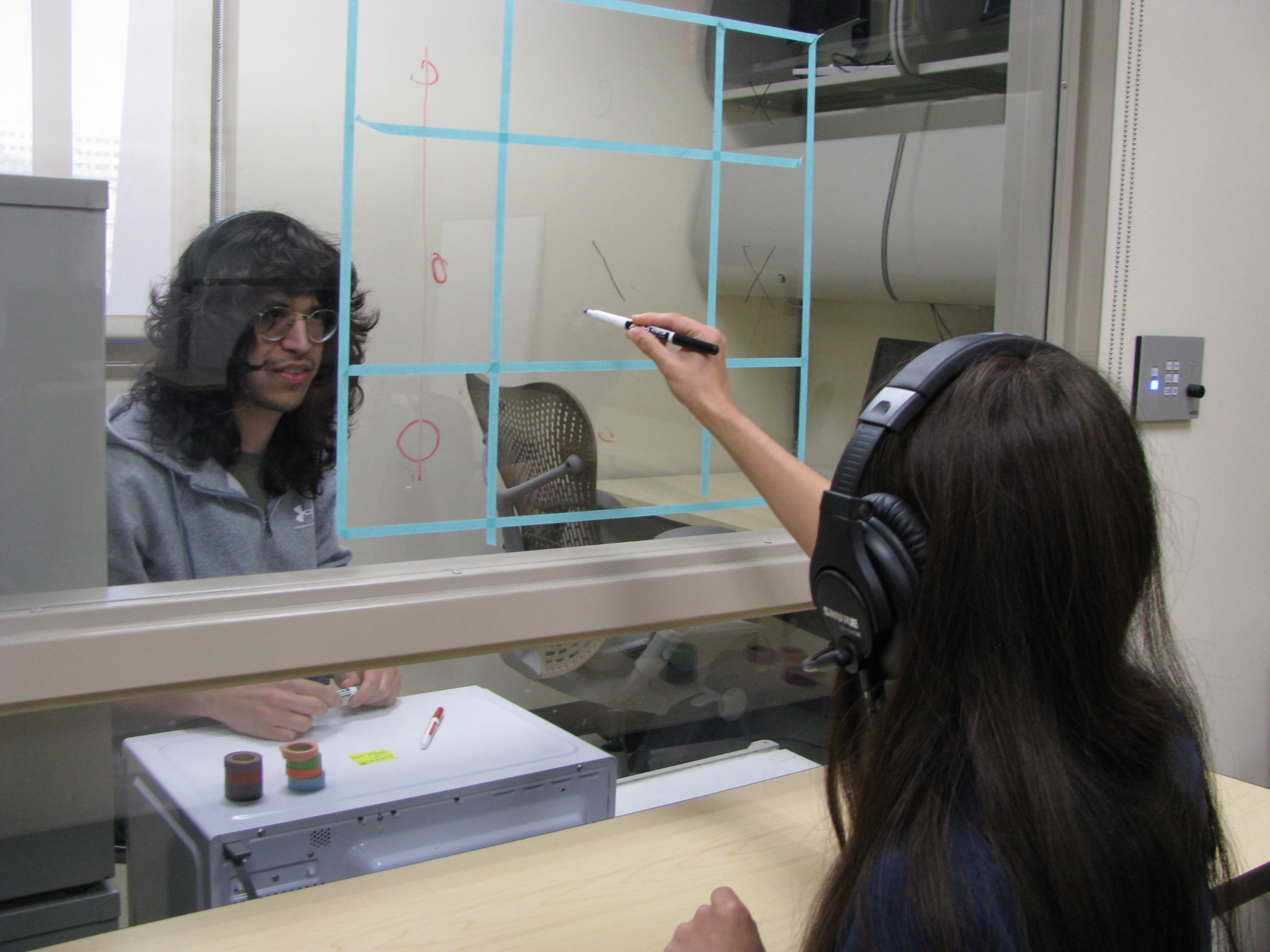}
\caption{Tic-tac-toe on the glass wall} \label{fig:tic-tac-toe}
\end{figure}

\item Play a game of tic-tac-toe.  As seen in Figure \ref{fig:tic-tac-toe}, 
he participants played games of tic-tac-toe with the tape or  markers. There were interesting utterances 
because they would often get competitive and react in different ways to the outcome of the game.
\item Try a snack. Talk about if you like it or dislike it and why.

When this prompt was given, the rooms were set up with a few snacks including different types of chocolate, lollipops, and a seaweed snack. The reactions of the participants varied. This produced utterances of disgust, confusion, satisfaction, and surprise. It also gave us rare utterances of people speaking while chewing.

\item What charity on the list do you think is the most deserving? If you two can reach agreement, tell me, and I’ll tell Professor Ward, who will give \$20 to the charity you pick. (He gives a few hundred dollars to El Paso charities every year anyway, so he’ll just reallocate \$20 based on your choice). PS. If you reach agreement quickly, please also come up with a one-sentence justification for your choice. You can either write this down, or just remember it to tell me later.

This prompt was intended to get utterances of negotiating, agreeing, and disagreeing. When the participants went through the process of choosing and justifying their choices, it resulted in a lot of discussion until they came to a consensus. This prompt led to participants talking about real world issues and dark topics which resulted in some emotional utterances.

\item You each have an envelope of quarters. You can 1) look for quarters that match, and if you and the other person find a pair, you can keep them, and/or 2) decide which two quarter designs are the best. If you need more light for this, feel free to turn on the light or raise the shades.

This prompt made participants choose between two options and often made them excited to try the first option. The conversations had a lot of back and forth as the participants tried to find a pair. They would describe the designs of their quarters and ask questions to the other participant to figure out what they had. The last suggestion was intended to elicit utterances marking separating from and then rejoining the conversation, but no participants went to change the lighting. 

\item  If you like, make yourself a paper-clip bracelet. After you and your conversation partner figure out the process, maybe go competitive, and see who can make one fastest.
  
Participants who tried this prompt would create the bracelet for themselves and talk about other topics while doing so. 
No participants took this competitively.

\item Can you tie up a second rubber band chain? Can you explain the process clearly enough for the other person to make one too?

For this activity, one participant got the rubber bands tied up in the specific manner while the other participant only got loose rubber bands. The participants would work together to figure out how to tie up the loose rubber bands in the same manner as the sample provided. Occasionally the participants would react when the rubber band would slip or move in a way they didn’t expect.

\item (Keep this suggestion from the other person.) A few seconds into the conversation, say the word “galafoos”. See how the other person reacts. Explain why you said it. Also, if you like, guess why we had you say a nonsense word, then afterwards ask the experimenter if you guessed right.

Participants that tried this prompt would confuse their conversation partner when they said “galafoos”. The confused participant would question what it meant and ask why it was said randomly. Sometimes they would create a small discussion to guess what the word could mean.

\item How much is the other person paying for their cell phone plan? How does it compare to yours? Would it make sense to change plans?

Conversations using this prompt would lead to participants explaining their cell phone plan and why they have it. Sometimes it would lead to participants talking about their financial situation. The last two questions of the prompt were mostly ignored.

\item Take these two partial maps, and the one who has the route marked, please explain it to the other so that they can draw it on their map.   

For this we use materials from the Maptask project 
\cite{maptask91}. The participant who received directions would sometimes be confused at certain points and would ask for clarification.

\item Abortion is illegal in Texas, as is helping someone else get an abortion. Are both of you okay with discussing this topic? Please remember that we will be publicly releasing these conversations, except for the parts you ask us to redact. If you’re not comfortable with this topic, talk about anything else. But if you are, consider the scenario where someone you know, maybe a cousin, is in a bad situation and comes to you desperate for help, needing a ride over to New Mexico for a few hours. How would you respond?

All participants who got this prompt decided not to discuss it.

\item Talk about whatever you like, maybe \{pet peeves, controversial opinions, TV shows or movies\}.

The pet peeves prompt resulted in utterances of annoyance and displeasure because participants would rant about their pet peeves. The controversial opinions prompt resulted in serious utterances. The topics of controversy that participants brought up were mostly about immigration and abortion. The TV shows and movies prompt had participants talking about plotlines and characters from the stories they, liked which made participants invested in the conversation.

\end{enumerate}

These suggestions were variously delivered verbally, on scraps of
paper, or in sealed envelopes not to be opened until in the recording rooms. 

Other prompts that we  considered but have not tried yet include: (a)
Work together to write out a plan for a brief tour of campus etc.\ for
a first-time visitor from Durango (or Arizona) who is considering
coming to UTEP for graduate school.  (b) Agree on a strategy for appealing the grading decisions of a hopelessly unfair TA. (c) Pick one of you to pretend to
want to travel by bus from UTEP to somewhere in the city (his/her
choice), and ask the other person to use their phone to find a way to
get there, and write down the route they find for you. (d) Have one
person describe how to get to his/her house from UTEP, while the other
takes notes well enough to get there without using an app. (e) Invent a game with this deck of cards that you can play through a glass wall. (f) Credit card debt can be a terrible thing. Ask the other person if they feel this way too, and if so, what strategies they have to stay out of trouble. Compare with your strategies.

The producer tended to make suggestions from the list above based on
her knowledge of what dialog activities had been rare recently, or
what she thought the participant pair would like.  It also turned out
that ad hoc topics suggested by the operator to appeal specifically to
the participants, for those whom she knew or came to know, could work
very well: for example ``whether or not it's a good idea to be in a
relationship while in college.''

\section{Cost Components }

We estimate that the total cost per matched pair of short fragments
was  between \$20 and \$30.

The major cost component is the time spent by the producer/operator/director/annotator.  (The other two are the stipends for the participants, and various ``overhead,'' including  post-processing, organization and quality control, and paperwork.)

Currently we are  spending about 75 producer hours per hour of yield, where an hour of yield means 1 hours worth of Spanish fragments matched with about 1 hours worth of English fragments.

This is based on the facts that our producer labor time for a 1-hour
recording session is almost 2 hours, and that over our first 28
sessions our yield was just over 45 minutes.  The breakdown: Each
session takes 5 minutes for the consent and demographic forms, 10
minutes for the initial conversation, and 45 minutes for the
re-enactments.  After that comes annotation of the data to identify
the matched pairs for our program to extract them.  This takes 15--45
minutes.  Most of this is in the second phase, where the annotator has
to carefully check that the phrases (short fragments) correspond
across the languages and carefully delimit them.  The time varies
based on factors such as the quality of the re-enactments: when this
is lower overall, the annotator has to be more judgmental and careful.

As noted earlier, we don't have a systematic quality control process;
rather we mostly rely on the producer to enforce quality as she goes.
However in real production, with a diverse set of producers, quality
control would almost certainly be necessary.  This would probably
require someone to rate each matched pair for quality, perhaps on a scale from
0 to 3, which might take 10 minutes, or more
if the quality control person is not the director, and therefore has
to listen to more context.  We'd then also need to modify our workflow
to copy over such ratings into the release.

The total time per session is thus 60 minutes for participant A, 60 minutes for participant B, and 120 minutes for the staff person.  Some efficiencies may be possible, but we feel that most of this time is probably not hugely squeezable.  In addition we have other staff time costs which are rather specific to our local procedure, including time for recruiting subjects, dealing with no-shows, setting up and breaking down the environment and equipment, and running the software to output the pairs.
Perhaps some of these could be avoided or better designed with a different setup or environment.  We note that these time-cost estimates exclude the start-up costs, which were huge, but should  be much less for future protocol modifications.

The next two sections discuss explorations with protocol alternatives, motivated  in
part by the goal of reducing cost.

\section{Other-Language Explorations}

This section describes experimental data collections in Japanese and Bengali.
While included in the release, these are experimental and intended to
illustrate what we can get, rather than to be used for modeling or evaluation.

Our main reason for collecting these was to explore the possibility of
scaling up to more languages without having to hire operators knowing each language.
To put it another way, a possible way to reduce costs would be to reduce the
qualifications required of the operator hired to handle the recording.
Our current protocol assumes that the
producer/director/operator/annotator is bilingual and thus fully able
to monitor quality in both languages, but it is possible to
increase the division of labor, such that the
director/operator needs only technical skills and people skills,  not
language skills.

Our second reason for doing these was to find any other issues
that might arise for language pairs beyond Spanish-English.

Accordingly we did two experimental data collections, one in English
and Japanese, and one in English and Bengali.  Our operator speaks
neither Japanese nor Bengali.  For Ja-En, one participant was the
first author, hence highly familiar with the protocol.  For Bn-En, one
participant was a lab member, hence broadly familiar with our aims and
methods.  All four participants in these experiments were less truly
bilingual than our average Spanish-English participant. We followed
the protocol, with two exceptions: in both sessions we tried both
directions, switching the source and target languages, and we omitted
the second annotation pass for the  phrase-level correspondences.
For both experiments, after processing was complete, we had a meeting
to review a sampling of the pairs collected, with each meeting
including one of the participants of the respective experiment.

The next subsection presents observations relating to
quality, the following subsection presents observations relating more
to language issues, and the last subsection summarizes the findings
regarding the possibility of recording with a language-unfamiliar operator.

\subsection{Quality-Related Observations}

Fidelity was often weaker than in the En-Es re-enactments.  This may
have been in part due to language differences making fidelity harder.
It may have also been due to reduced effortfulness.  In the Ja-En, we
observed a couple of cases where the participants failed to re-create
all the nuances, for example, treating something as new information in
the original utterance, but not in the re-enactment.  The ability of
the director to assess fidelity and request retakes was very limited,
although in one case she was able to suggest a retake after noticing
that the re-enactment was suspiciously shorter and more fluent than
the original.

Regarding naturalness, there were cases where aspects of the prosody
of Japanese leaked into the English re-enactment.  There appeared to
be more of these than in the Es-En.  This may have been partly due to
a language skill deficit.

In terms of throughput, we were concerned that it would take more time
to identify interesting and complete utterances to re-enact.  However,
the director was able to estimate when an utterance started and ended
by using the sound wave display in Elan, and most candidate utterances
that she selected on this basis were confirmed by the participants to
be natural units to re-enact.  Overall the rate of data collection was
good, however there were two confounding factors: having a
knowledgeable participant and having apparently lower quality
aspirations.

Regarding annotation quality: Here the usual operator did the
annotation, based on her timepoint notes from the re-enactment phase.
This went well, except in one case where there was an error in her
notes which, not knowing the language, she was unable to later correct
for.

\subsection{Language-Related Observations}

As these were our first excursions beyond English and Spanish, we
noted some other issues, which likely relate more to the language
pairs than to the operator's knowledge.  These observations are
impressionistic and should be revisited through systematic analysis.

First, regarding both languages, we observed that word order
differences between Bn-En and Ja-En seem more extreme than between
Es-En.  This would make it harder to identify sub-utterance phrases
(short fragments) that match up across the two languages.

Regarding Japanese: (1) Japanese and English are so different in
structure and that it was at times impossible to translate fully
faithfully. (2) Japanese is more interactive than English, so many
behaviors, such as backchanelling, overlapped speech, and laughter,
were often simply missing in the English re-enactments.  (3) Japanese
is less explicit than English, so sometimes the English came out
significantly longer.

Regarding Bengali: (1) The Bengali re-enactments were often
significantly longer, which may reflect a language or cultural
tendency to speak in full sentences.  (2) Prosodically, we noticed that
Bengali seemed to differ from English in the expressions of pausing
for a moment to think of a word, of listing things, and of marking
questions.  (3) Grammatically, Bengali has different word order, with
the verb frequently at the end, and with prepositional phrases
sometimes fronted.

Other quality-related comments: (1) The Bengali data included several
quite long utterances, which the participants sometimes seemed to
struggle to re-enact.  This may have been in part because the topic
chosen was ``my research.''  While our confederate had suggested
lighter topics, such as movies, the other participant had declined
these.  (2) There were times where the re-enactment included thinking
of the Bengali word while speaking.  This led to gratuitous
mismatches, which the operator couldn't catch. A possible mitigation
to this specific problem could be to add an instruction to
``completely formulate the utterance in your mind, before you start
speaking,'' perhaps augmented with ``this can be hard, so feel free to
practice once or twice until you're ready.''

\subsection{Conclusion and Possible Mitigations}

Overall, having a language-unfamiliar operator clearly reduces the
quality obtained.  In comparison to Es-En, where perhaps 95--98\% of
the pairs were fully acceptable, for Jp-En this was probably closer to
75\%. Of course, the low quality pairs could still have some utility
as part of a large collection.  For Bn-En the yield was probably even
lower, with some of the low-quality pairs so poor as to be likely of
no value to any currently conceivable machine learning scheme.

This quality deficit occurred despite a situation that was nearly
ideal in some respects: most of the participants were motivated and
highly cooperative, and the operator/director was experienced,
motivated, engaged, and sensitive enough to prosody to notice likely
lapses in fidelity or naturalness.  Despite this, the quality was
frequently poor.  The major factor is probably whether or not the
operator/director  can judge quality, apply ``peer pressure,'' and
request retakes.  Without this, even for motivated participants,
the quality will suffer.

If it is nevertheless necessary to record with a language-unfamiliar
operator, there may be ways to partly mitigate the quality loss, including:

Mitigation via procedure modification: One might periodically nag
the participants about the goal of fidelity and  occasionally ask
them to rate the quality of their re-enactments; and similarly for
the goal of naturalness.

Mitigation via participant selection: With a language-unfamiliar
director, it becomes more important to have participants who are
highly cooperative, sensitive to nuance, and diligent.

Mitigation via using experienced participants: Our confederate later
participated in another Bn-En session, and reported that her
experience in the first session helped her become a better guide.
Thus, it could help to ensure that at least one participant in each
pair has experience with the protocol.

Mitigation via participant training: We may also wish to produce a
more polished, formal description of what we want, perhaps in the form
of a video.  This could stress our goals and provide illustrations of
what we consider good re-enactments and of common ``error'' types
(too literal, slavishly similar prosody, etc.).  We might even back this up with a follow-on
quiz, and only consider people qualified to participate if they pass
it. We could go further and require participants to be
indoctrinated and trained in our aims and procedure, including perhaps
having participated in quality review of a dozen utterance pairs, so
that they truly understand what we want.

Mitigation in the annotation phase: While our operator was able to
produce utterance-level (long-fragment) annotations, to obtain
phrase-level (short-fragment) annotations would require hiring an
annotator who knows both languages.

Mitigation via a quality-control phase: In our standard protocol the
quality is ensured during the re-enactment phase.  If the operator is
language-unfamiliar, one mitigation would be to add a formal
quality-control phase, to identify and probably discard low-quality
utterances.

As all of these mitigations would increase the time cost, and likely
recover only some of the lost quality.  We conclude that using a
language-familiar operator is likely still the best strategy overall.

\section{An Exploration in Remote Collection}

Currently our recordings are done in-lab, which requires participants
to come to us.  On campus this is not a major problem, but to scale up,
it would be nice to develop a protocol for remote data collection.
This section discusses the issues expected and those observed,
based in part on a small experimental
data collection using Zoom, also described.

\subsection{Concerns}

A priori, based on our general knowledge of recording methods and of remote
conversations, many things could go wrong in remote recording.

First, there are some basic issues of audio quality:
\begin{enumerate}
      \item Participants' connectivity issues may hurt audio quality or introduce delay.  (observed)
      \item Participants' audio environments will vary, due to room reverberation,
            background noise levels, transient noise, etc. (observed)
      \item The audio quality will be poorer, since we cannot provide high-quality
            microphones. (observed)
      \item The recorded audio quality in the interaction may be poorer, due
            to the videochat software's compression and packet-loss concealment algorithms. (observed)
      \item Transmission delay may make it impossible, depending on the videochat software,
            to easily  synch up the different tracks. (intermittently observed)
\end{enumerate}

Second, there are considerations of impact on the conversations;
remote conversations will be different from the face-to-face ones.
Given that our experiment was mostly about getting the procedure to work,
we did not specifically confirm whether or not these were happening.
\begin{enumerate}
      \item Altered dialog dynamics, due to transmission delay --- both audio and video, with the
            former mattering more --- that affect the conversation dynamics of
            the original conversation, and maybe also limiting what can be
            re-enacted. Previous research suggests that the most severe
            effect will be on the turn-taking, but there is a silver lining
            in that reduced overlap, laughter and backchanneling will make
            post-processing and modeling easier. In this way,  remote data collection may have
            the benefit of excluding data that would be hard to model, and
            which is not representative of any real use case.
      \item Altered speech and interactive behaviors, due to limited video bandwidth,
            such as seeing the face only, without hand gestures etc.
      \item Altered behavior due to lack of a shared environment and shared visual
            references.
      \item Reduced ability for pre-dialog synching of mood, which is likely to make the
            dialogs more formal. This could be mitigated in various ways, but a more
            formal style may actually better serve some use cases.
      \item Reduced ability to reinforce expectations, which may reduce the extent to
            which the participants take the re-enactment task seriously, which may reduce both
            the naturalness and fidelity of the pairs collected.
\end{enumerate}

Third, there may be implications for participant behavior that affect process efficiency:
\begin{enumerate}

      \item Channel lag and limited bandwidth interaction may reduce
        the ability of the director to deftly and swiftly guide
        re-enactments, making it a slower process.  (observed)
      \item Remote effects may reduce the enjoyability of the process.  (observed)
      \item Remote collection may increase the cognitive load on the director. (observed)

\end{enumerate}

So, there are numerous  quality-cost trade-offs and time-cost  trade-offs
to consider. 

Incidentally, while we here have been assumed that the director and
the participants will all be remote from each other, there is also the
possibility of allowing, or encouraging, the participants to be
co-located.  This would hurt audio separation, and decrease the
likelihood of getting participants who are strangers to each other,
which is important for the main use case.  But it could be easier to
recruit subjects and perhaps more enjoyable for them.

The rest of this section describes the process that we developed,
in part to offer one remote workflow, and in part to describe the
context in which we made the above observations.

\subsection{Remote Collection Preparation using Zoom}

The subsection describes the procedure we used in our pilot experiment.
There are probably better tools for doing remote collections.

\subsubsection{Install and Configure Zoom}

\begin{enumerate}
      \item Install the Zoom desktop client, version 2.0 or higher.
      \item Open the Zoom desktop client and sign in. Create an account if you do not
            already have one.
      \item Click the user account icon on the top right, then click ``Settings''.
      \item In the ``Settings'' window, click the ``General'' tab.
      \item Enable ``Show my meeting duration''.
      \item In the ``Settings'' window, click the ``Recording'' tab.
      \item Note the location meeting recordings will be stored to, next to ``Store my
            recordings at:''. This folder will open automatically after ending a
            meeting.
      \item Enable ``Record a separate audio file for each participant''.
      \item Close the ``Settings'' window.
      \item (macOS) Open System Settings, then click the ``Privacy \& Security'' tab.
            Under ``Screen Recording'', enable screen recording for the Zoom
            application.
\end{enumerate}

\subsubsection{Install and Configure ELAN}

\begin{figure}[ht]
      \centering
      \includegraphics[scale=0.5]{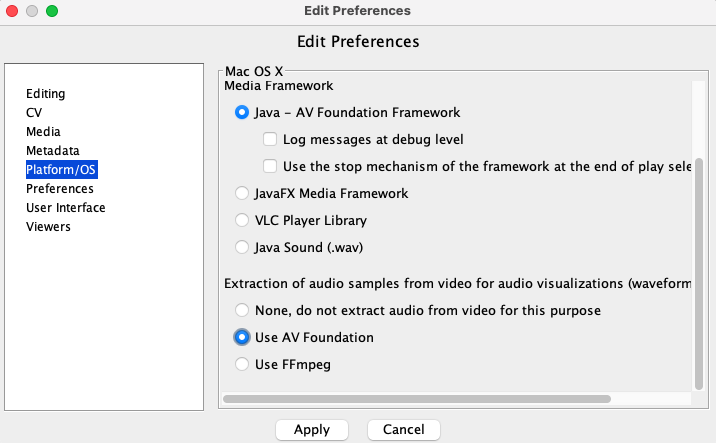}
      \caption{ELAN ``Platform/OS'' settings after changing the ``Extraction of
            audio\ldots'' setting to ``Use AV
            Foundation''.}\label{fig:elan-platform-os-settings}
\end{figure}

\begin{figure}[ht]
      \centering
      \includegraphics[scale=0.5]{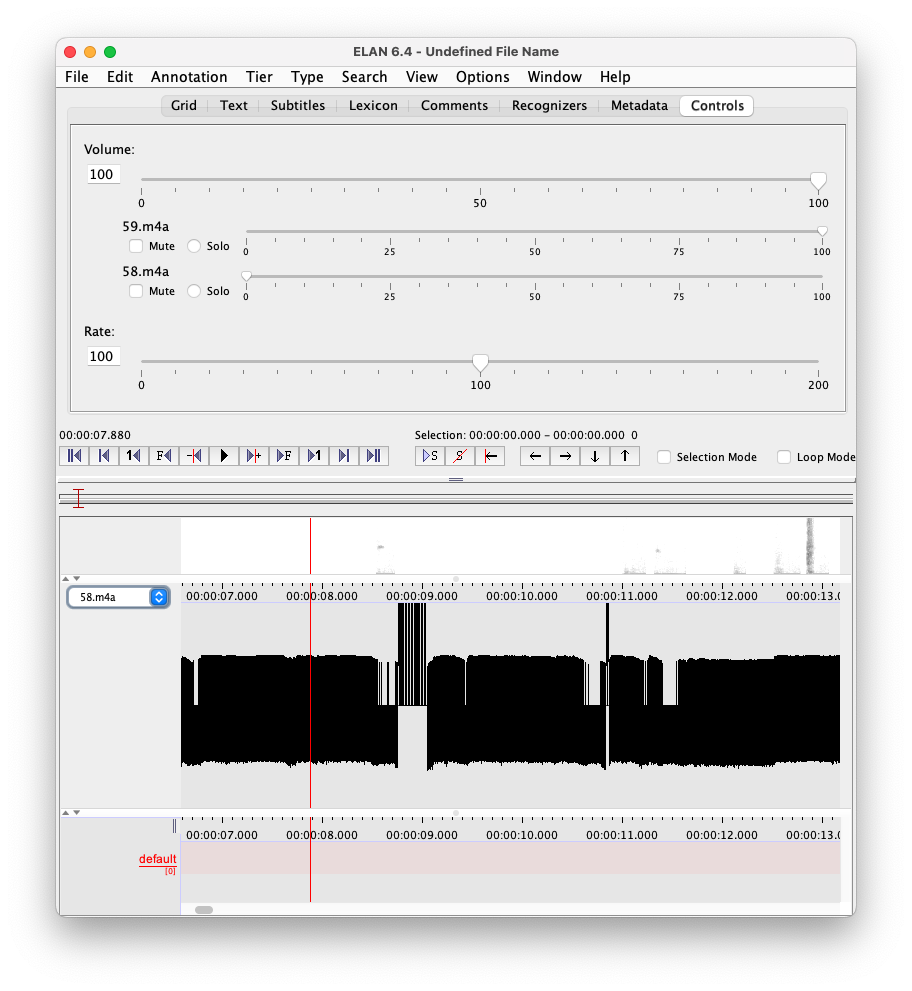}
      \caption{ELAN waveform visualization using the default ``Use FFmpeg''
            setting.}\label{fig:elan-ffmpeg-setting}
\end{figure}

\begin{figure}[ht]
      \centering
      \includegraphics[scale=0.5]{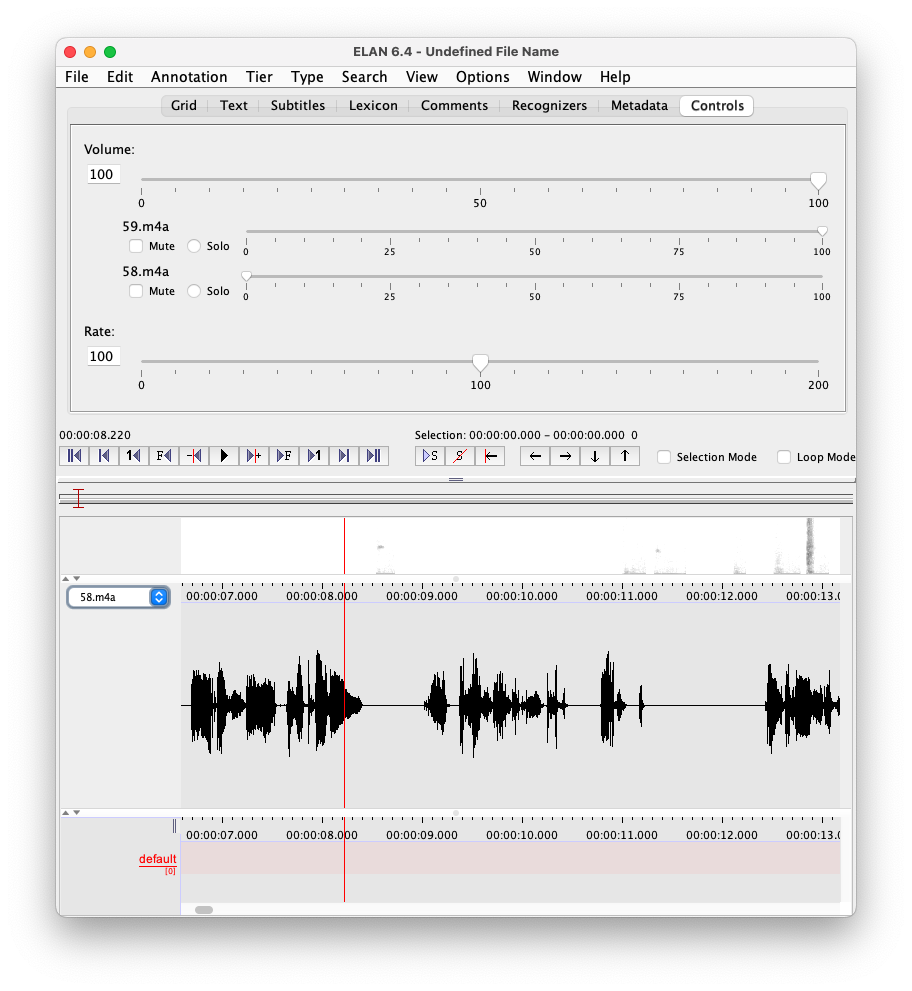}
      \caption{ELAN waveform visualization using the ``Use AV Foundation''
            setting.}\label{fig:elan-av-setting}
\end{figure}

\begin{enumerate}
      \item Install ELAN, version 6.4 or higher.
      \item The default settings may result in strange-looking waveform visualizations.
            This may be specific to the audio coding format (MPEG-4) or may be specific
            to macOS.\@ To fix the visualizations, click ``Preferences\ldots'', then
            click the ``Platform/OS'' tab, and under ``Extraction for audio samples from
            video for audio visualizations (waveform, spectrogram)'' select ``Use AV
            Foundation'' (Figure~\ref{fig:elan-platform-os-settings}). Compare the
            waveform visualization before (Figure~\ref{fig:elan-ffmpeg-setting}) and
            after (Figure~\ref{fig:elan-av-setting}) changing this setting.
\end{enumerate}

Also create a folder to store all conversation audio and markup files, e.g. ``DRAL''.

\subsubsection{Prepare to record the original conversation}
\begin{enumerate}
      \item Collect the participants' email addresses.
      \item As in the in-person procedure, enter a new row for each participant in the
            ``participant'' metadata sheet.
\end{enumerate}

\subsubsection{Record the original conversation}\label{sec:record-original-conv}
\begin{figure}[ht]
      \centering
      \includegraphics[scale=0.5]{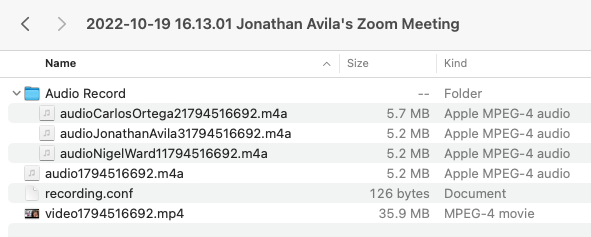}
      \caption{Zoom recording files contents.}\label{fig:recorded-meeting-contents}
\end{figure}

\begin{enumerate}
      \item Start a Zoom meeting as the host.
      \item Invite the participants to the meeting. Click ``Participants'', then click
            ``Invite'', then click the ``Email'' tab, then send an email invitation to
            the participants.
      \item Once the participants join the meeting, click ``Record''.
      \item After the participants end their conversation, let them know they will take
            a break and join a second meeting shortly.
      \item End the meeting. Zoom will begin converting the recording.
      \item Once the recording has been processed, the folder containing the recording
            files will open.
      \item Within the recording folder, open the ``Audio Record'' folder. In the
            ``Audio Record'' folder, each participant's recorded audio will be listed as
            its own audio file (.m4a), with a filename containing the name they used in
            the meeting (Figure~\ref{fig:recorded-meeting-contents}).

\end{enumerate}

\subsubsection{Prepare to record the re-enactments}

\begin{figure}[ht]
      \centering
      \includegraphics[scale=0.5]{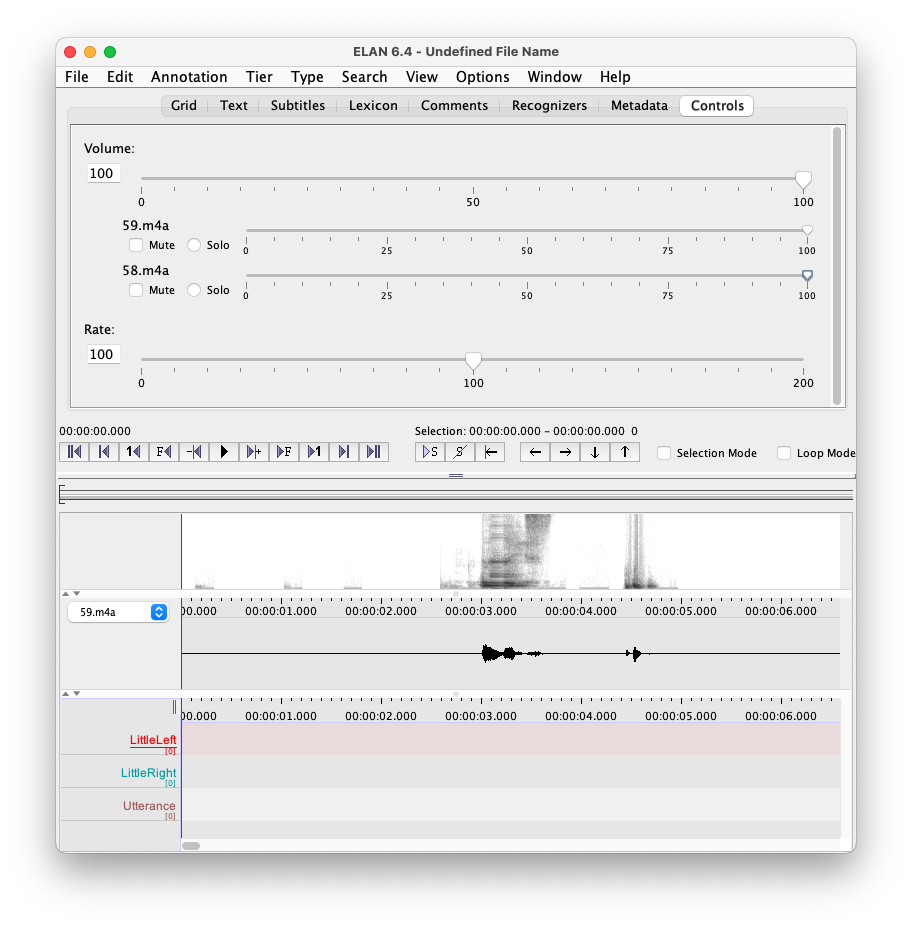}
      \caption{ELAN main window after importing audios, adjusting volume, and
            configuring annotation tiers.}\label{fig:elan-ready-to-save}
\end{figure}

\begin{figure}[ht]
      \centering
      \includegraphics[scale=0.5]{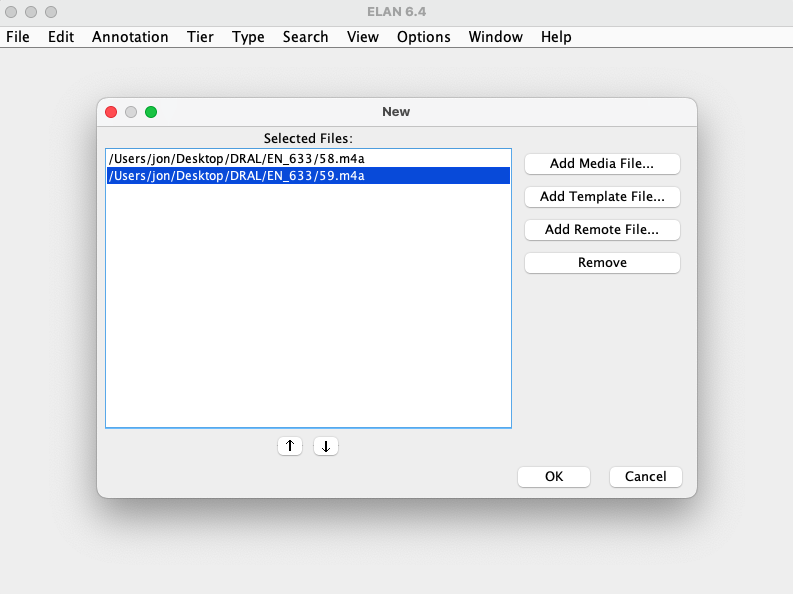}
      \caption{ELAN ``New'' window after importing participant
            audios.}\label{fig:elan-importing-audio-files}
\end{figure}

\begin{enumerate}
    \item Copy both participants' audio files to the ``DRAL'' folder.
      \item Rename each participant's audio file with their participant ID
            and language code, e.g. ``61-ES.m4a''. The language code is added to avoid duplicate filenames when adding the audio files from the re-enactment conversation later.
        
      \item Delete the Zoom meeting recording folder with the remaining files: the audio
            of the operator, the combined meeting audio, the recording configuration
            file, and the meeting video.
      \item Open ELAN.\@
      \item Click ``File'', then click ``New\ldots''.
            \item Import the participants' audio files. In the ``New'' window, click ``Add
            Media File\ldots'', then browse to select the
            audio of the first participant, then click ``Open''. Repeat for the second
            audio.
      \item After selecting both audio files
            (Figure~\ref{fig:elan-importing-audio-files}), click ``OK''.
      \item By default, ELAN will set the volume of the first audio to 100\% and the
            volume of all other audios to 0\%. Click the ``Controls'' tab, then raise
            the volume of the second audio to 100\%.
      \item Configure the annotation tiers, like in the in-person procedure.
      \item After adjusting the volume and configuring the tiers, the ELAN window should
            look like Figure~\ref{fig:elan-ready-to-save}.
            \item Click ``File'', then click ``Save'' and save the markup file (.eaf) in the
            ``DRAL'' folder, with the same name as the conversation ID.\@
\end{enumerate}

\begin{figure}[ht]
      \centering
      \includegraphics[scale=0.5]{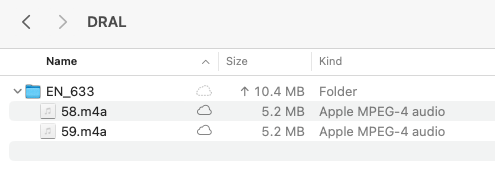}
      \caption{Conversation folder contents before opening
            ELAN.}\label{fig:file-structure-before-elan}
\end{figure}

\subsubsection{Record the re-enactments}

\begin{enumerate}
      \item Start a Zoom meeting as the host, invite the participants to the meeting,
            and start recording the meeting, as in
            Section~\ref{sec:record-original-conv}. Note the time into the meeting when
            the recording starts.
      \item Share the ELAN window with the original conversation audios so that you and
            the participants can hear its audio. Click ``Share Screen'', then select
            ``ELAN \textendash{} ELAN [version number] \textendash{} [ELAN filename]'',
            then click ``Share''.
            \begin{itemize}
                  \item Note: The ELAN waveform display may appear glitchy after sharing
                        your screen. We were not able to resolve this issue in our tests.
            \end{itemize}
      \item As in the in-person procedure, select the fragments to re-enact and note
            down the times of the re-enactments, using the time-into-the-meeting
            displayed by Zoom.
            \begin{itemize}
                  \item Note: The time into the meeting displayed by Zoom is not the
                        same as the time into the recording. Use the time into the meeting
                        when the recording started you noted in Step 1 above to adjust
                        accordingly.
                  \item Note: When using multiple audio files, ELAN displays just one
                        waveform at a time. To change the waveform display to a different
                        audio, select the audio from the drop-down list to the left of the
                        waveform display.
                  \item Note: The two audios may sound out of sync at some times. We
                        were not able to resolve this issue in our tests.
                  \item Note: The audio recorded and processed by Zoom is exactly what
                        was heard during the meeting. Any problems due to lag or interruptions
                        during this phase should be fixed by re-re-enacting.
            \end{itemize}

      \item End the meeting and wait for the folder containing the recording files to
            open, as in Section~\ref{sec:record-original-conv}.
\end{enumerate}

\subsubsection{Markup the conversation fragment pairs}

\begin{enumerate}
      \item Copy both participant's audio files to the ``DRAL'' folder and rename
            each participant's audio file with their participant ID and language code, as in
            Section~\ref{sec:record-original-conv}.
      \item Create a new ELAN file, import the two audios, adjust the volume and
            configure the annotation tiers, and save the markup file (.eaf) with the
            conversation ID, as in Section~\ref{sec:record-original-conv}.
      \item Like in the in-person procedure, use your notes to annotate the conversation
            fragment pairs.
      \item Like in the in-person procedure, enter a new row for the original
            conversation in the ``conversation'' metadata sheet and enter another row
            for the re-enactment conversation.
\end{enumerate}

\subsubsection{Prepare the data for post-processing}

\begin{enumerate}

    \item After completing the data collection, the file structure of the ``DRAL''
            folder should be one flat folder with both Elan files and four audio files.  
    \item Convert the Zoom audio recordings from M4A format to WAV, e.g. with the FFmpeg command: \texttt{ffmpeg -i path-to-input.m4a path-to-output.wav}.
    \begin{itemize}
        \item To install FFmpeg, see instructions at: \url{https://ffmpeg.org/download.html}
        \item Depending on your operating system and install method, you may need to add FFmpeg to your PATH environment variable to be able to execute it from anywhere in the command line. Alternatively, replace the `ffmpeg` in the command above with the path to FFmpeg above.
    \end{itemize}
    \item Merge the two mono audio files into one stereo audio file with the SoX command: \texttt{sox -M path-to-input-left.wav path-to-input-right.wav path-to-output.wav}. Like the in-person procedure, name the output audio with the matching conversation ID.
    \begin{itemize}
        \item Instructions for installing SoX are in Section~\ref{post-processing-setup}.
    \end{itemize}
    \item Continue the post-processing procedure in Section~\ref{post-processing-procedure}.
    \item Archive the ``DRAL'' folder with \texttt{tar} or another archiving utility,
            e.g. \\
            \texttt{tar -vcf
                  \textquotesingle{}\textquotesingle{}DRAL-2.0.tgz\textquotesingle{}\textquotesingle{}
                  [path to DRAL folder]}
\end{enumerate}


\bigskip\noindent {\bf Acknowledgments:} We thank Ann Lee, Justine
Kao, Carleigh Wood, Benjamin Peloquin, and Ines Boglioli for
discussion and suggestions. This work was supported in part by a University Research Institute grant from the University of Texas at El Paso. 

\printbibliography{}

\clearpage

\appendix

\section{Examples from the corpus}\label{appendix:examples}

Below are examples from the corpus. Inferrable elided continuations are in brackets.
Matched phrases are in bold.

\subsection{Example 1 {\normalfont(ES\_008\_18,
                  EN\_008\_18)}}\label{appendix:example-mismatch}

\noindent Original: \textit{¿Van a venir, venir tus papás para \ldots}

\noindent Gloss:  will-they to come, come your parents to \\
\hspace*{3.6em}[help you move in]?

\noindent Re-enactment: \textit{Are you parents gonna come, or} \\
\hspace*{3.6em}\textit{[not]?}

\noindent Prosody Note: both original and re-enactment end with a saliently lengthened
syllable, seemly inviting the listener to complete the phrase

\medskip \noindent In context:

\smallskip
\noindent
\begin{tabular}{lp{30em}} X:\@ & \textit{Vas a tener tu propio,}                         \\
               Y:\@            & \textit{Ai, si cierto.}                                 \\
               X:\@            & \textit{departamento.}                                  \\
               Y:\@            & \textit{Ya el jueves.}                                  \\
               X:\@            & \textit{¿El jueves?}                                    \\
               Y:\@            & \textit{El jueves me lo van a dar, el jueves a las tres
               de la tarde.}                                                             \\
               X:\@            & \textit{\textbf{¿Van a venir, venir tus papás para?}}
\end{tabular}

\medskip \noindent Translation of context:

\smallskip\noindent
\begin{tabular}{lp{40em}} X:\@ & \textit{You're going to have your own,}                \\
               Y:\@            & \textit{Ah, that's right.}                             \\
               X:\@            & \textit{apartment.}                                    \\
               Y:\@            & \textit{Already on Thursday.}                          \\
               X:\@            & \textit{On Thursday?}                                  \\
               Y:\@            & \textit{On Thursday they're going to give it to me, on
               Thursday at three in the afternoon.}                                     \\
               X:\@            & \textit{\textbf{Are you parents gonna come, or?}}
\end{tabular}

\subsection{Example 2 \normalfont{(EN\_006\_29,
            ES\_006\_29)}}\label{appendix:example-prosody}

Context: X spent some time working with inmates at the county jail

\noindent Original, in context:

\smallskip\noindent
\begin{tabular}{lp{40em}} X:\@ & \textit{So I worked over there, it was pretty\ldots
               \textbf{interesting}.}                                                \\
               Y:\@            & (laughter)                                          \\
               X:\@            & \textit{It was, it was a cool experience.}          \\
               Y:\@            & (overlapping) \textit{Good choice of words.}        \\
\end{tabular}

\noindent Prosody Note: The word \textit{interesting} came after a 2.0 s pause.

\noindent Inferred Pragmatic Intent: indicate that there were things that he doesn't
want to talk about

\medskip\noindent Re-enactment:

\noindent
\begin{tabular}{lp{17em}} X:\@ & \textbf{\it \bf interesante.}
\end{tabular}


\subsection{Example 3 \normalfont{(ES\_002\_3.34,
            EN\_002\_4.34)}}\label{appendix:example-spontaneous}

\noindent Original: {\it Lo estoy interpretando como yo quiero y como yo pienso.}

\noindent Gloss: It I-am construing as I wish and \\
\hspace*{3.6em} as I am-thinking.

\noindent Inferred Pragmatic Intent: assert his intention to tell his story without
being censored

\noindent Prosody Note: both original and re-enactment have extra punch on the stressed
syllables, giving an impression of annoyance and assertiveness

\noindent Re-enactment: {\it I am interpreting things how I want and how I think.}

\medskip\noindent In context:

\smallskip\noindent
\begin{tabular}{lp{36em}} X:\@ & \textit{Todo que es incómodo a trabajar en la tierra, a
               tumbar becerros,}                                                         \\
               Y:\@            & (overlapping) \textit{A sí.}                            \\
               X:\@            & \textit{cortarle los \ldots}                            \\
               Y:\@            & (gasp)                                                  \\
               X:\@            & \textit{Eso viene siendo cosas que a muchas personas
               los incomoda, y yo ya estoy acostumbrado hacerlo.}                        \\
               Y:\@            & \textit{Sabes que me acabo de acordar que este audio lo
                     vamos a, a distribuir, osea, entero, así que van a oír todo lo que
               dijimos desde el principio.}                                              \\
               X:\@            & \textit{Pero pues yo estoy diciendo cosas. \textbf{Lo
                           estoy interpretando como yo quiero y como yo pienso.}}
               \\
               Y:\@            & \textit{A ok, bueno pues, perdón.}
\end{tabular}

\medskip\noindent Translation of context:

\smallskip\noindent
\begin{tabular}{lp{36em}} X:\@ & \textit{Everything that is uncomfortable when working
               on the land, throwing down calves,}                                     \\
               Y:\@            & (overlapping) \textit{Oh yes.}                        \\
               X:\@            & \textit{cutting off their \ldots}                     \\
               Y:\@            & (gasp)                                                \\
               X:\@            & \textit{Those are things that make many people
               uncomfortable, and I'm used to doing it.}                               \\
               Y:\@            & \textit{You know, I just remembered that this audio
                     we're going to, to distribute, like, whole, so they're going to
               hear everything we said from the beginning.}                            \\
               X:\@            & \textit{Well, I'm saying things. \textbf{I am
               interpreting things how I want and how I think.}}                       \\
               Y:\@            & \textit{Ah okay, well then, sorry.}
\end{tabular}

\vspace{5cm}
\section{Language/dialect background form}\label{appendix:background-form}

\includepdf[pages=-,pagecommand={},width=\textwidth]{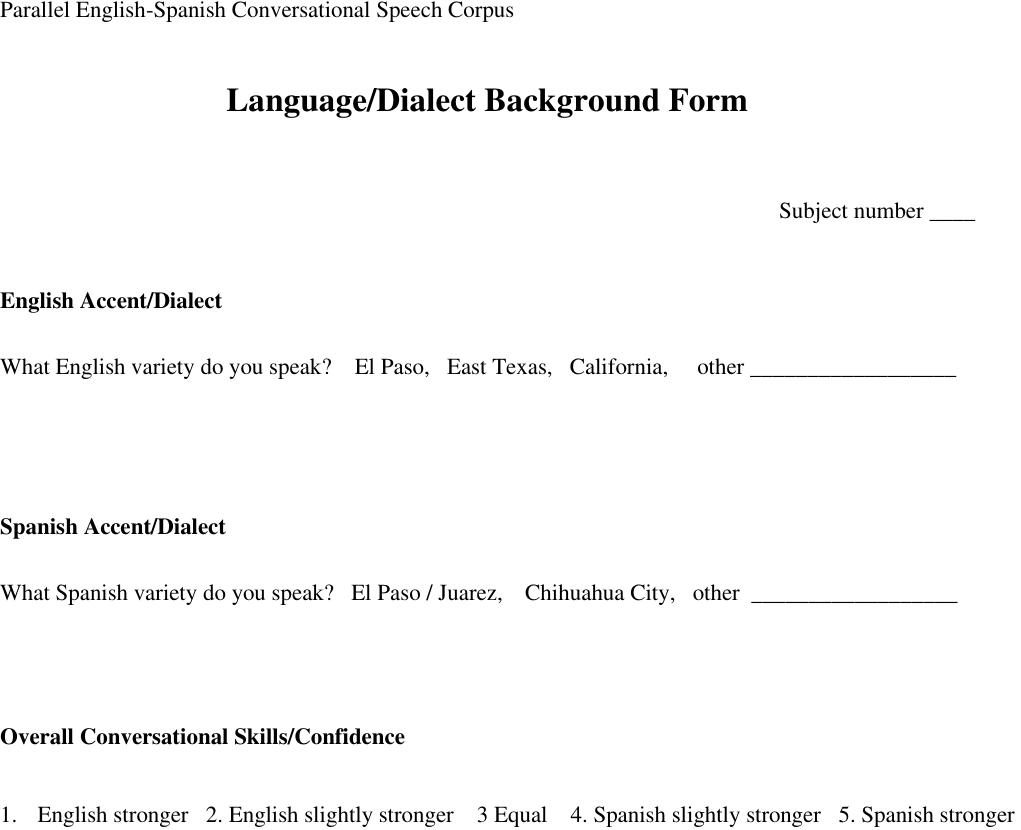}

\section{Advertisement}\label{appendix:advertisement}

\includepdf[pages=-,pagecommand={},width=15cm]{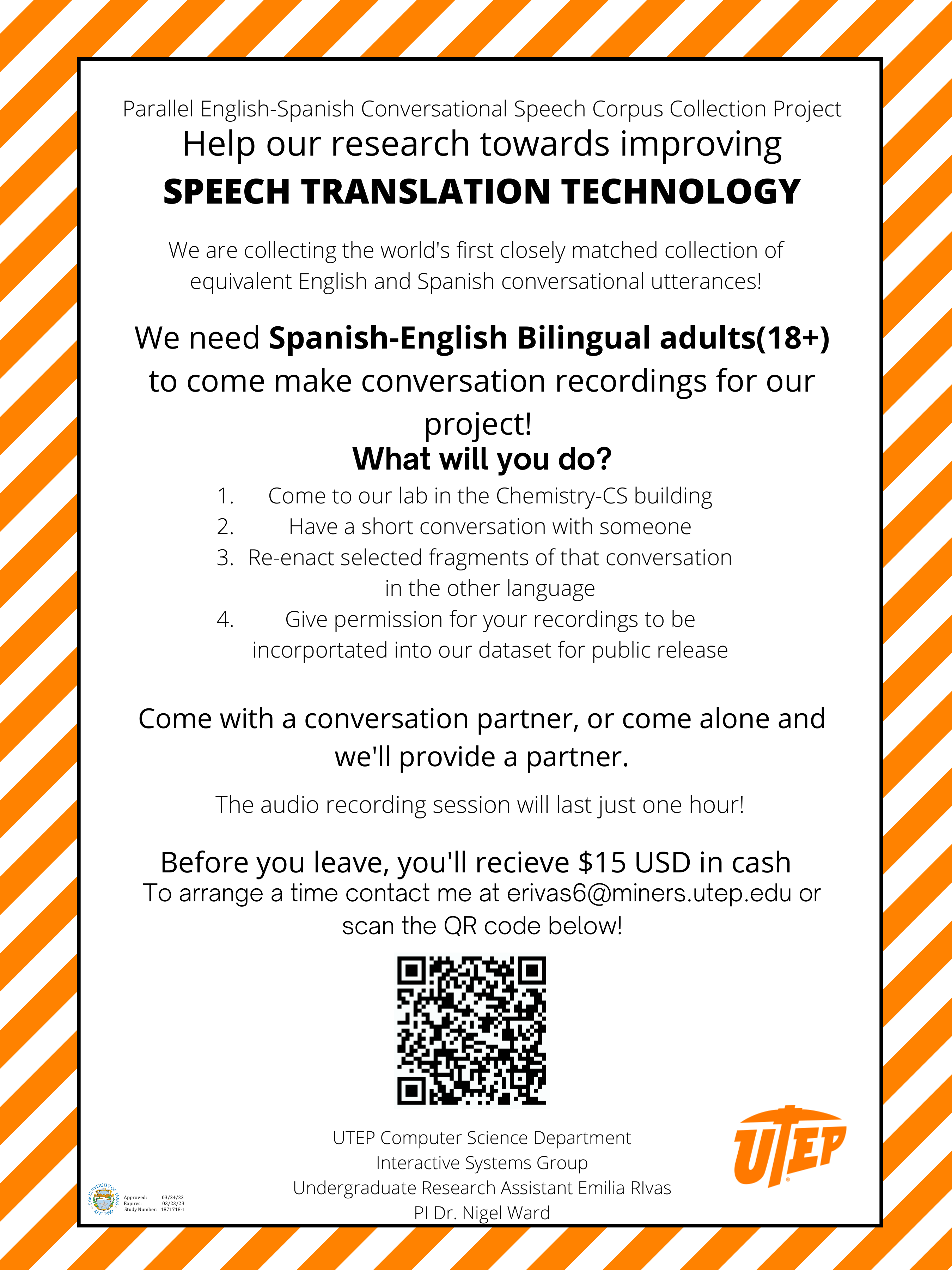}

\section{Informed consent form}\label{appendix:consent-form}

\includepdf[pages=-,pagecommand={},width=\textwidth]{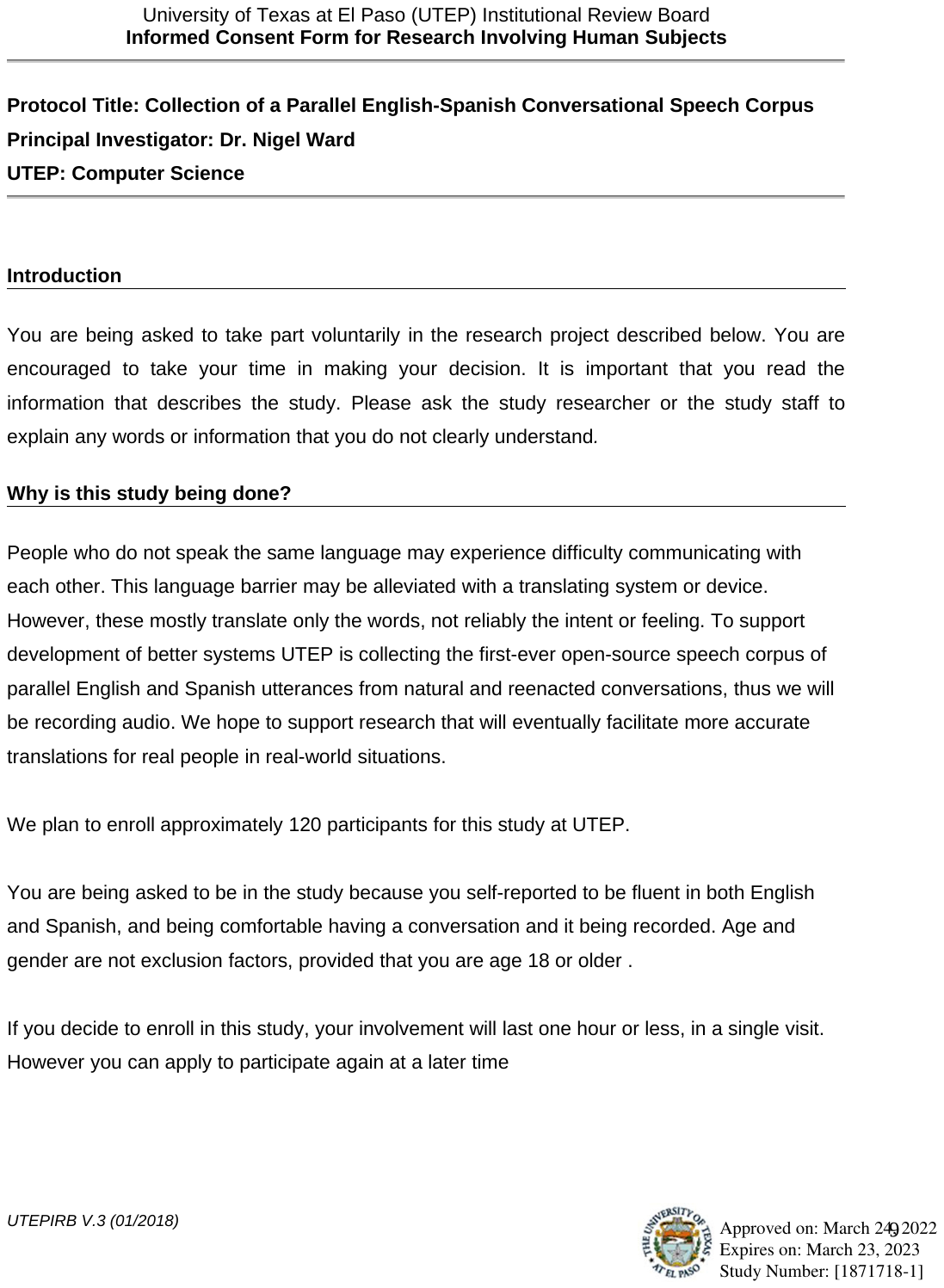}

\section{ELAN keyboard shortcuts}\label{appendix:elan-keyboard-shortcuts}
\begin{itemize}
      \item To highlight a span of the audio, left-click and drag the cursor across
            the span.
      \item CTRL + Space: Play / Pause
      \item Shift + Space: Play / Pause highlighted area
      \item CTRL + Alt + N:\@ A new annotation is created for the selected and
            highlighted span.
      \item CTRL + Alt + \(<\) or \(>\): Selects the previous or next annotated span.
\end{itemize}

\end{document}